\documentclass[12pt,runningheads]{llncs}

\usepackage{graphicx}
\usepackage{amsmath,amssymb}

\usepackage{geometry}
\DeclareMathOperator{\argmax}{argmax}
\DeclareMathOperator{\softmax}{softmax}
\usepackage{caption, subcaption}
\newtheorem{thm}{Theorem}
\newtheorem{cor}[thm]{Corollary}
\newtheorem{lem}[thm]{Lemma}

\newtheorem{defn}[thm]{Definition}

\newtheorem{exmpl}[thm]{Example}
\newtheorem{alg}[thm]{Algorithm}

\usepackage{hhline}
\usepackage{helvet}
\usepackage{multicol}
\usepackage{courier}
\usepackage{color}
\usepackage{bbm}
\usepackage{enumerate}

\usepackage{caption, subcaption}
\begin{document}

\title{Generalization Error in Deep Learning}

\titlerunning{Generalization Error in Deep Learning}

\author{Daniel Jakubovitz \inst{1} \and
Raja Giryes\inst{1} \and
Miguel R. D. Rodrigues\inst{2}}

\authorrunning{D. Jakubovitz, R. Giryes, M. R. D. Rodrigues}

\institute{School of Electrical Engineering,\\
	Tel Aviv University, Israel\\
	\email{danielshaij@mail.tau.ac.il, raja@tauex.tau.ac.il} \and
Department of Electronic and Electrical Engineering,\\ University College London, UK\\
\email{m.rodrigues@ucl.ac.uk}
}

\maketitle

\abstract{Deep learning models have lately shown great performance in various fields such as computer vision, speech recognition, speech translation, and natural language processing.
However, alongside their state-of-the-art performance, it is still generally unclear what is the source of their generalization ability. Thus, an important question is what makes deep neural networks able to generalize well from the training set to new data. In this article, we provide an overview of the existing theory and bounds for the characterization of the generalization error of deep neural networks, combining both classical and more recent theoretical and empirical results.}

\section{Introduction}
\label{Introduction}
Deep neural networks (DNNs) have lately shown tremendous empirical performance in many applications in various fields such as computer vision, speech recognition, speech translation, and natural language processing \cite{Goodfellow16Deep}. However, alongside their state-of-the-art performance in these domains, the source of their success and the reason for their being a powerful machine learning model remains elusive.

A deep neural network is a complex non-linear model, whose training involves the solution of a non-convex optimization problem, usually solved with some variation of the stochastic gradient descent (SGD) algorithm. Even though convergence to a minimum with good performance is not guaranteed, it is often the case that the training of DNNs achieves both a small training error and good generalization results.

This article focuses on the characterization of the generalization abilities of neural networks. Indeed, there are various recent theoretical advances that aim to shed light on the performance of deep neural networks, borrowing from optimization theory, approximation theory, and related fields (e.g. see \cite{Global_Optimality_in_Neural_Network_Training,MathematicsOfDeepLearning} and others).
Yet, due to space constraints, we concentrate here on over-viewing recent prominent approaches of statistical learning theory for understanding the generalization of deep neural networks.

The generalization error of a machine learning model is the difference between the empirical loss of the training set and the expected loss of a test set. In practice, it is measured by the difference between the error of the training data and the one of the test data. This measure represents the ability of the trained model (algorithm) to generalize well from the learning data to new unseen data.
It is typically understood that good generalization is obtained when a machine learning model does not memorize the training data, but rather learns some underlying rule associated with the data generation process, thereby being able to extrapolate that rule from the training data to new unseen data and generalize well.

Therefore, the generalization error of DNNs has been the focus of extensive research, mainly aimed at better understanding the source of their capabilities and deriving key rules and relations between a network's architecture, the used optimization algorithm for training and the network's performance on a designated task. Bounds on the generalization error of deep learning models have also been obtained, typically under specific constraints (e.g. a bound for a two-layer neural network with ReLU activations).
Recent research also focuses on new techniques for reducing a network's generalization error, increasing its stability to input data variability and increasing its robustness.

The capabilities of deep learning models are often examined under notions of expressivity and capacity: their ability to learn a function of some complexity from a given set of examples. It has been shown that deep learning models are capable of high expressivity, and are hence able to learn any function under certain architectural constraints.
However, classical measures of machine learning model expressivity (such as Vapnik-Chervonenkis (VC) dimension \cite{VC_dimension}, Rademacher complexity \cite{RademacherBounds}, etc.), which successfully characterize the behavior of many machine learning algorithms, fail to explain the generalization abilities of DNNs.
Since DNNs are typically over-parameterized models with substantially less training data than model parameters, they are expected to over-fit the training data and obtain poor generalization as a consequence \cite{RethinkingGeneralization}. However, this is not the case in practice. Thus, a specific line of work has been dedicated to study the generalization of these networks.

Several different theories have been suggested to explain what makes a DNN generalize well. As a result, several different bounds for the generalization error of DNNs have been proposed along with techniques for obtaining better generalization in practice.
These rely on measures such as the PAC-Bayes theory \cite{Pac-Bayes_Framework,Pac-Bayes_Framework_2,Simplified_PAC_Bayesian_Margin_Bounds}, algorithm stability \cite{Stability_and_Generalization},  algorithm robustness \cite{RobustnessGeneralization} and more.

In the following sections we survey the theoretical foundations of the generalization capabilities of machine learning models with a specific emphasis on deep neural networks, the corresponding bounds on their generalization error, and several insights and techniques for reducing this error in practice. Namely, we review both classical and more recent theoretical works and empirical findings related to the generalization capabilities of machine learning algorithms, and specifically deep neural networks.

\section{The learning problem}
\label{The learning problem}

Machine learning is a field which employs statistical models in order to \emph{learn} how to perform a designated task without having to explicitly program for it.
It is closely related to and inspired by other domains in applied mathematics, computer science and engineering such as optimization, data mining, pattern recognition, statistics and more. Accordingly, machine learning models and methods are inspired by the prominent techniques, models and algorithms of these fields \cite{Goodfellow16Deep,Murphy2013Machine}.

Core to the field of machine learning is the learning (training) process, in which an algorithm is given a training dataset in order to learn how to perform a desired task by learning some underlying rule associated with the data generation process. 
After the learning phase is done, a good algorithm is expected to perform its task well on unseen data drawn from the same underlying rule. This phase is commonly referred to as the test phase, in which an algorithm performs its designated task on new data. In general, machine learning can be divided into two categories. The first category is supervised learning, in which there is a ground truth value (label) for each data sample. This ground truth value is supplied to the algorithm as part of its training dataset, and is expected to be correctly predicted by the algorithm during the test phase for new unseen data. The second category is unsupervised learning, in which there are no ground truth labels that characterize the data, and it is up to the algorithm itself to characterize the data correctly and efficiently in order to perform its task.

Some of the most prominent machine learning algorithms are the Support Vector Machine (SVM), $K$-Nearest Neighbors ($K$-NN), $K$-Means, decision trees, deep neural networks etc. \cite{Murphy2013Machine}. These algorithms are used to perform a variety of different tasks such as regression, classification, clustering and more. Deep neural networks, which are the subject of this article, are a particular model (algorithm) that has attracted much interest in the past several years due to its astonishing performance and generalization capabilities in a variety of tasks \cite{Goodfellow16Deep}.

The following notation is used throughout this article.
The input space of a learning algorithm is the $D$-dimensional subspace $\mathcal{X} \subseteq{\mathbb{R}^D}$ and $x \in \mathcal{X}$ is an input sample to the algorithm. The output space is the $K$-dimensional subspace $\mathcal{Y} \subseteq{\mathbb{R}^K}$. The label of the input sample $x$ is $y \in \mathcal{Y}$. The sample set is denoted as $\mathcal{Z}$, where $\mathcal{Z} = \mathcal{Y} \times \mathcal{X}$.

We will concentrate predominantly on classification tasks. Consequently, $K$ corresponds to the number of possible classes and $k^* \in \{1,\dots,K\}$ is the correct class of the input sample $x$.
Accordingly, the label vector $y \in \mathcal{Y}$ is a one-hot vector, meaning all its elements are equal to zero except for the element in the $k^*$-th index which is equal to one.
The function $f_W$ is the learned function of a model with parameters $W \in \mathcal{W}$ (where $\mathcal{W}$ is the parameter space), i.e. $f_W: \mathbb{R}^D \rightarrow \mathbb{R}^K$.
The $k$-th value of the vector $f_W(x)$ is denoted by $f_W(x)[k]$. In most cases we omit the $W$ symbolizing the network's parameters for convenience.

We assume a training dataset has $N$ training examples, such that the set $\mathbf{s}_N =  \left\{ s_i |  s_i \in \mathcal{Z} \right\}_{i=1}^{N} = \left\{ (x_i,y_i) \in \mathcal{X} \times \mathcal{Y} \right\}_{i=1}^{N}$ is the algorithm's training set.
The samples are independently drawn from a probability distribution $\mathcal{D}$.
The set $\mathbf{t}_{N_{test}} = \left\{ t_i | t_i \in \mathcal{Z} \right\}_{i=1}^{N_{test}} =  \left\{ (x_i,y_i) \in \mathcal{X} \times \mathcal{Y} \right\}_{i=1}^{N_{test}}$ is the algorithm's test set, which consists of $N_{test}$ test examples.
The hypothesis set, which consists of all possible functions $f_W$, is denoted as $\mathcal{H}$. Therefore, a learning algorithm $\mathcal{A}$ is a mapping from $\mathcal{Z}^{N}$ to $\mathcal{H}$, i.e. $\mathcal{A}: \mathcal{Z}^{N} \rightarrow \mathcal{H}$.

The loss function, which measures the discrepancy between the true label $y$ and the algorithm's estimated label $f(x)$ is denoted by $\ell \left( y,f(x) \right)$. In some cases, when referring to the loss of a specific learning algorithm $\mathcal{A}$ trained on the set $\mathbf{s}_N$ and evaluated on the sample $s$, we denote $\ell (\mathcal{A}_{\mathbf{s}_N}, s)$ instead.

For a general loss function $\ell$, an algorithm's empirical loss on the training set (train loss) $\mathbf{s}_N$ is
\begin{align}
\ell_{emp}(f,\mathbf{s}_N) = \ell_{emp}(\mathcal{A}_{\mathbf{s}_N})
\triangleq \frac{1}{N} \sum_{i=1}^{N} \ell \left( y_i,f(x_i) \right), \hspace{3mm} \{(x_i,y_i)\}_{i=1}^{N} \in \mathbf{s}_{N},
\end{align}
and the expected loss of the algorithm is
\begin{align}
\ell_{exp}(f) = \ell_{exp}(\mathcal{A}_{\mathbf{s}_N}) \triangleq \mathbb{E}_{(x,y) \sim \mathcal{D}} \left[ \ell (y,f(x)) \right].
\end{align}
Accordingly, an algorithm's \emph{generalization error} is given by
\begin{align}
GE(f,\mathbf{s}_N) \triangleq \left| \ell_{emp}(f,\mathbf{s}_N)  - \ell_{exp}(f) \right|.
\end{align}
The empirical test loss is often used to approximate the expected loss since the distribution $\mathcal{D}$ is unknown to the learning algorithm. 
The test loss of an algorithm is given by
\begin{align}
\ell_{test}(f,\mathbf{t}_{N_{test}}) \triangleq \frac{1}{N_{test}} \sum_{i=1}^{N_{test}} \ell (y_i,f(x_i)), \hspace{3mm} \{(x_i,y_i)\}_{i=1}^{N_{test}} \in \mathbf{t}_{N_{test}}, 
\end{align}
and the corresponding approximation of the \emph{generalization error} is given by
\begin{align}
GE(f,\mathbf{s}_N,\mathbf{t}_{N_{test}}) \triangleq \left| \ell_{emp}(f,\mathbf{s}_N)  - \ell_{test}(f,\mathbf{t}_{N_{test}}) \right|.
\end{align}

The output classification margin $\gamma$ of a data sample $x$ is defined by the difference between the value of the correct class and the maximal value over all other classes: $\gamma = f(x)[k^*] - \max_{k \neq k^*} f(x)[k]$, where as mentioned earlier, $k^*$ corresponds to the index associated with the correct class of the data sample $x$, i.e. $y[k^*]=1$ and $y[k] = 0$ $\forall k \neq k^*$. The margin loss is defined as follows.

The empirical margin loss for an output margin $\gamma$ is
\begin{align}
{\ell}_{emp,\gamma} (f,\mathbf{s}_N) \triangleq \frac{1}{N} \sum_{i=1}^{N} \mathbbm{1} \left\{ f(x_i)[k^*_i] - \max_{k \neq k^*_i}f(x_i)[k] \leq \gamma \right\},
\hspace{1mm} {\{(x_i,y_i)\}_{i=1}^{N} \in \mathbf{s}_N},
\end{align}
where $\mathbbm{1}$ signifies the indicator function that gets the value one if the inequality holds and the value zero otherwise.
The expected margin loss for an output margin $\gamma$ is
\begin{align}
{\ell}_{exp,\gamma}(f) \triangleq Pr_{(x,y) \sim \mathcal{D}} \left[ f(x)[k^*] -\max_{k \neq k^*}f(x)[k] \leq \gamma \right].
\end{align}

We denote the Frobenius, $\ell_1$, $\ell_2$ and $\ell_{\infty}$ norms by $||\cdot||_F$, $||\cdot||_1$, $||\cdot||_2$ and $||\cdot||_{\infty}$ respectively.

The training of machine learning algorithms relies on the Empirical Risk Minimization (ERM) principle. Since a learning algorithm only has access to a finite amount of samples drawn from the probability distribution $\mathcal{D}$, which is unknown, it aims at minimizing the \emph{empirical risk} represented by the training loss $\ell_{emp}(\mathcal{A}_{\mathbf{s}_N})$. This practice can be sub-optimal, as it is subject to the risk of over-fitting the specific training samples. The term "over-fitting" refers to a phenomenon in which a learning algorithm fits the specifics of the training samples "too well", thereby representing the underlying distribution $\mathcal{D}$ poorly.

Throughout this article the notion of model (algorithm) \emph{capacity} is used. This is a general term that relates to the capability of a model to represent functions of a certain complexity. It is evaluated in different ways in different contexts. A formal and accurate definition is given where relevant along this article.

Two classical metrics which are used to evaluate the capacity (or expressivity) of learning algorithms are the VC-dimension \cite{VC_dimension} and the Rademacher complexity \cite{RademacherBounds}.  
The VC-dimension measures the classification capacity of a set of learned functions.

\begin{defn}(VC-dimension)
A classification function $f$ with parameters $W$ shatters a set of data samples $\left\{ x_i \right\}_{i=1}^{N}$ if for all possible corresponding labels $\left\{ y_i \right\}_{i=1}^{N}$ there exist parameters $W$ such that $f$ makes no classification errors on this set.
The VC-dimension of $f$ is the maximum amount of data samples $N$ such that $f$ shatters the set $\left\{ x_i \right\}_{i=1}^{N}$. If no such maximal value exists then the VC-dimension is equal to infinity.
\end{defn}

\begin{figure}
\centering
\begin{subfigure}[b]{0.45\linewidth}
\centering
\includegraphics[width=40mm]{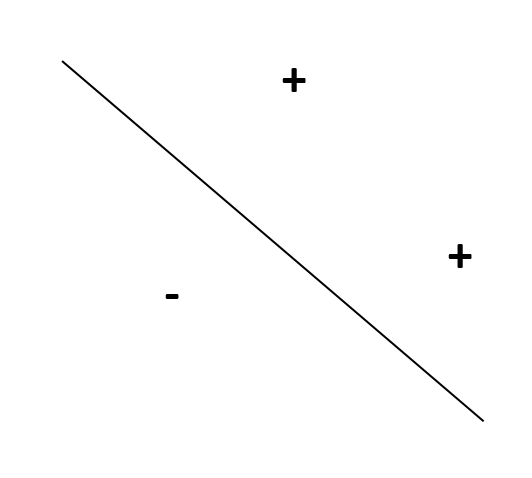}
\caption{Three samples in $\mathbb{R}^2$ correctly classified by a linear function.}
\label{VCdim_a}
\end{subfigure}
\begin{subfigure}[b]{0.45\linewidth}
\centering
\includegraphics[width=40mm]{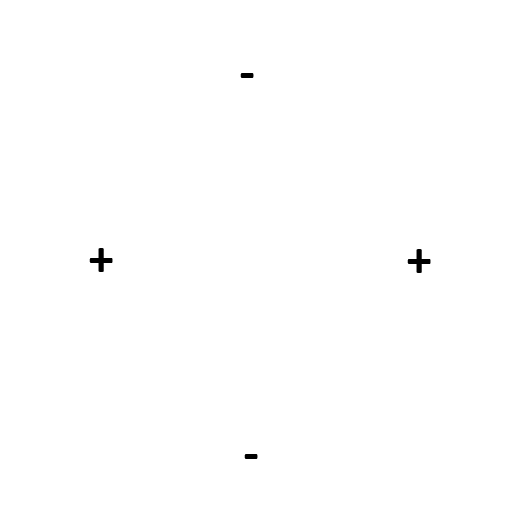}
\caption{Four samples in $\mathbb{R}^2$ which cannot be correctly classified by a linear function.}
\label{VCdim_b}
\end{subfigure}
\caption{Linear classification of samples in $\mathbb{R}^2$.}
\label{VCdim_illustration}
\end{figure}

To gain intuition as to the meaning of the VC-dimension, let us consider the following example.
\begin{exmpl}
Let us consider a linear function in the space $\mathbb{R}^2$. The function $\alpha_1 x_1 + \alpha_2 x_2 + b = 0$, which is parameterized by $W=(\alpha_1, \alpha_2, b) \in \mathbb{R}^3$, defines a classification decision for any sample $x = (x_1,x_2) \in \mathbb{R}^2$ according to the following rule:
\begin{equation}
  f(x)=\begin{cases}
    +1, & \text{if $\alpha_1 x_1 + \alpha_2 x_2 + b \geq 0$}\\
    -1, & \text{otherwise}
  \end{cases}
\end{equation}
This means that any sample above or on the line is classified as positive ($+1$), whereas any sample under the line is classified as negative ($-1$).
Let us define the hypothesis set:
\begin{align}
\mathcal{H} = \{ f_W | W = (\alpha_1, \alpha_2, b) \in \mathbb{R}^3 \},
\end{align}
then $\mathrm{VCdim}(\mathcal{H}) = 3$.
\end{exmpl}
\textbf{Proof sketch.} Note that in this case any three samples in $\mathbb{R}^2$ (which are not co-linear) can be shattered by a linear classifier. However, four samples in $\mathbb{R}^2$ can be easily chosen such that no linear function can represent the correct classification rule.
See Fig.~\ref{VCdim_illustration} for an illustration of these cases.

The Rademacher complexity measures the richness of a set of functions with respect to some probability distribution. Essentially, it measures the ability of a set of functions to fit random $\pm 1$ labels. 

\begin{defn}(Rademacher complexity)
Given a dataset $\mathbf{s_x} = \{ (x_i) \}_{i=1}^{N}$, and a hypothesis set of functions $\mathcal{H}$, the empirical Rademacher complexity of $\mathcal{H}$ given $\mathbf{s_x}$ is 
\begin{align}
\label{Rademacher complexity}
{\mathcal{R}}_{N}(\mathcal{H}) = \mathbb{E}_{\sigma} \left[ \sup_{h \in \mathcal{H}} \frac{1}{N} \sum_{i=1}^{N} \sigma_i h(x_i)  \right],
\end{align}
where $\sigma_{i} \in \left\{ \pm 1 \right\}, i=1, \dots, N$ are independent and identically distributed uniform random variables, i.e. $Pr(\sigma_i = 1) = Pr(\sigma_i = -1) = \frac{1}{2}, i=1,\dots,N$.
\end{defn}
Note that since $h(x_i) \in \{\pm 1\}$, if $\sigma_i h(x_i) = 1$ the classification is correct and if $\sigma_i h(x_i) = -1$ the classification is wrong, and therefore we seek to maximize the sum in \eqref{Rademacher complexity}.
To gain intuition as to the meaning of the Rademacher complexity, let us consider the following example of linear classifiers.

\begin{exmpl}
Let $\mathcal{H}$ be the class of linear classifiers with an $\ell_2$ norm of the weights bounded by $\alpha$: $\{ \{ w^T x_i \}_{i=1}^{N}, ||w||_2 \leq \alpha \}$. Let us assume that the $\ell_2$ norm of the samples in the dataset $\mathbf{s_x} = \{ x_i \}_{i=1}^{N}$ is upper bounded by $\beta$, i.e. $||x_i||_2 \leq \beta, \forall i=1,\dots,N$. Then the Rademacher complexity of $\mathcal{H}$ is upper bounded as
\begin{align}
\mathcal{R}_N (\mathcal{H}) \leq \frac{\alpha \beta}{\sqrt{N}}.
\end{align}
\end{exmpl}
\noindent \textbf{Proof.}
According to the definition of the Rademacher complexity,
\begin{align}
{\mathcal{R}}_{N}(\mathcal{H})
=
\mathbb{E}_{\sigma} \left[ \sup_{w: ||w||_2 \leq \alpha} \frac{1}{N} \sum_{i=1}^{N} \sigma_i w^T x_i  \right]
\end{align}
\begin{align}
=
\frac{1}{N} \mathbb{E}_{\sigma} \left[ \sup_{w: ||w||_2 \leq \alpha} w^T \left(\sum_{i=1}^{N} \sigma_i x_i \right)  \right]
\leq
\frac{\alpha}{N} \mathbb{E}_{\sigma} \left[|| \sum_{i=1}^{N} \sigma_i x_i ||_2 \right]
\end{align}
\begin{align}
\label{Jensen}
\leq
\frac{\alpha}{N} \sqrt{\mathbb{E}_{\sigma} \left[|| \sum_{i=1}^{N} \sigma_i x_i ||_2^2 \right]}
=
\frac{\alpha}{N} \sqrt{\mathbb{E}_{\sigma} \left[|| \sum_{i=1}^{N} x_i ||_2^2 \right]}
\end{align}
\begin{align}
=
\frac{\alpha \beta}{\sqrt{N}},
\end{align}
where in \eqref{Jensen} we used Jensen's inequality.

As described throughout this article, bounds on these measures of complexity of a learned function are generally unable to explain the generalization capabilities of deep neural networks, and are therefore more suited for the analysis of classical, less complex machine learning algorithms such as the support vector machines (SVM), $K$-NN and others \cite{Murphy2013Machine,Understanding_Machine_Learning}.

Another commonly used framework for the analysis of machine learning algorithms is the PAC-Bayes theorem (Probably Approximately Correct), which is used to bound the generalization error of stochastic classifiers \cite{Pac-Bayes_Framework,Pac-Bayes_Framework_2,Simplified_PAC_Bayesian_Margin_Bounds}. The PAC-Bayes framework provides a generalization bound which relates a prior distribution $P$, postulated before any data was seen, and a posterior distribution $Q$, which depends on the data (i.e. the training set). Unlike the VC-dimension and Rademacher complexity, the PAC-Bayes framework refers to the \emph{distribution} of the hypothesis set of learned functions rather than a specific classification function. One should keep in mind that this is a general framework for the analysis of machine learning algorithms, from which several different bounds and mathematical formulations have been derived. We present hereafter one of its core theorems.

\begin{thm}(PAC-Bayes theorem)
Let $\mathcal{D}$ be a distribution over $\mathcal{Y} \times \mathcal{X}$ from which examples are drawn independently.
Let $P$ and $Q$ denote probability distributions over the hypothesis set of classifiers $\mathcal{H}$.
In addition, let $Err_{\mathcal{D}}(Q) = \mathbb{E}_{f \sim Q} [Pr_{(x,y) \sim \mathcal{D}} (y \neq f(x))]$ be the expected test probability of error and let $Err_{\mathbf{s}}(Q) = \frac{1}{N} \sum_{i=1}^{N} \mathbb{E}_{f \sim Q} \left[ \mathbbm{1} \{ y_i \neq f(x_i) \} \right], \{ (x_i,y_i) \}_{i=1}^N \in \mathbf{s}_N$ be the expected empirical training probability of error of the stochastic classifier $Q$ over the training set $\mathbf{s}_N$ sampled from $\mathcal{D}$.
Then for any $Q$ and any $\delta \in (0,1)$, we have with probability at least $1-\delta$ that over $N$ randomly drawn training samples,
\begin{align}
KL \left( Err_{\mathbf{s}}(Q) || Err_{\mathcal{D}}(Q) \right) \leq \frac{KL(Q||P) + \ln(\frac{N+1}{\delta})}{N} 
\end{align}
holds for all distributions $P$.
\end{thm}

In the above, $KL(\cdot||\cdot)$ denotes the Kullback-Leibler divergence between two probability distributions and $\mathbbm{1}$ denotes the indicator function which gets the value one if the inequality holds and zero otherwise. Note that $P$ is an a-priori distribution and $Q$ is a posterior distribution given the training dataset $\mathbf{s}_N$.

The notion of algorithm robustness was introduced in \cite{RobustnessGeneralization}.
A learning algorithm is said to be \emph{robust} if for a training sample and a test sample that are close to each other, a similar performance is achieved. The following is the formal definition of a robust learning algorithm.

\begin{defn} (Robustness)
\label{Robustness definition}
Algorithm $\mathcal{A}$ is $(K,\epsilon(\mathbf{s}))$ robust if $\mathcal{Z}$ can be partitioned into $K$ disjoint sets, denoted as $\left\{ C_i \right\}_{i=1}^{K}$, such that $\forall s \in \mathbf{s}$,
\begin{align}
\label{Robustness_Definition}
s, z \in C_i, \Rightarrow |\ell(\mathcal{A_{\mathbf{s}}},s)-\ell(\mathcal{A_{\mathbf{s}}},z)| \leq \epsilon(\mathbf{s}).
\end{align}
\end{defn}
Note that $\ell(\mathcal{A_{\mathbf{s}}},s)$ is the loss on the sample $s$ of the algorithm $\mathcal{A_\mathbf{s}}$ which was trained on the set $\mathbf{s}$.
A weaker definition of robustness, pseudo-robustness, is also useful for the analysis of the generalization error of learning algorithms, and is given in subsection~\ref{Robustness and Generalization}.

The notion of \emph{sharpness} of the obtained solution to the minimization problem of the training of DNNs, i.e. the minimizer of the training loss, has lately become key in the analysis of the generalization capabilities of DNNs. Though several different definitions exist, we rely on the definition from \cite{Large_Batch_Training} which is in wide use. Formally, the sharpness of the obtained minimizer is determined by the eigenvalues of the Hessian matrix $\nabla^2 \ell_{emp}(\mathcal{A}_{\mathbf{s}_N})$ evaluated at the minimizer. However, since the computation of the Hessian matrix of DNNs is computationally expensive, an alternative measure is used.
This measure relies on the evaluation of the maximal value of $\nabla^2 \ell_{emp}(\mathcal{A}_{\mathbf{s}_N})$ in the environment of the examined solution. The maximization is done both on the entire input space $\mathbb{R}^D$ and on $P$-dimensional random manifolds, using a random matrix $A_{D \times P}$. 

\begin{defn} (Sharpness)
\label{Sharpness_definition}
Let $C_{\epsilon}$ denote a box around the solution over which the maximization of $\ell$ is performed. The constraint $C_{\epsilon}$ is defined by
\begin{align}
C_{\epsilon} = \left\{ z \in \mathbb{R}^{{P}} : -\epsilon (|(A^{\dagger}x)_i|+1) \leq z_i \leq \epsilon (|(A^{\dagger}x)_i|+1) \hspace{3mm}  \forall i \in \{ 1,\dots,{P} \} \right\}
\end{align}
where $A^{\dagger}$ denotes the pseudo-inverse of $A$. The value of $\epsilon > 0$ controls the size of the box.
Given $x \in \mathbb{R}^D$, $\epsilon > 0$ and $A \in \mathbb{R}^{D \times P}$, the $(C_{\epsilon},A)$-sharpness of $\ell$ at $x$ is defined by
\begin{align}
\phi_{x,\ell}(\epsilon,A) \triangleq \frac{(\max_{y \in C_{\epsilon}} \ell(x+Ay))-\ell(x)}{1+\ell(x)} \times 100.
\end{align}
\end{defn}

In a recent work \cite{StrongerBoundsCompression}, a compression based approach is used to derive bounds on the generalization error of a classifier. A compressible classifier is defined as follows.

\begin{defn} (Compressibility)
Let $f$ be a classifier and $G_{\mathcal{W}} = \{ g_W | W \in \mathcal{W} \}$ be a class of classifiers such that $g_W$ is uniquely determined by $W$. Then $f$ is $(\gamma,\mathbf{s})$-compressible via $G_{\mathcal{W}}$ with an output margin $\gamma > 0$, 
if there exists $W \in \mathcal{W}$ such that for any sample in the dataset $x\in \mathbf{s}$
%with a label $y \in \mathbb{R}^K$,
%such that $y[k^*]=1$, $y[k]=0$ $\forall k \neq k^*$,
we have for all $k$
\begin{align}
|f(x)[k] - g_W(x)[k] | \leq \gamma.
\end{align}
\end{defn}
Note that $f(x)[k]$ is the $k$-th entry in the $K$-dimensional vector $f(x)$.

\section{Deep neural networks}
\label{Deep neural networks}

In this section, we give the definition of a deep neural network and explain several aspects of its architecture. Readers familiar with deep neural networks can skip directly to section~\ref{Generalization Error in Deep Learning}.

\begin{figure}[b]
\centering
\includegraphics[scale=.5]{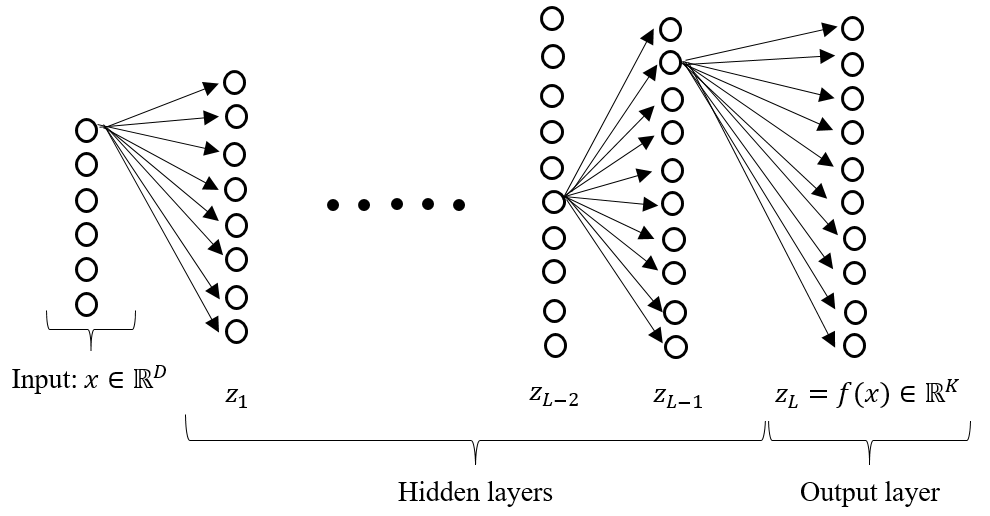}
\caption{A deep neural network. Some of the connections are omitted for simplicity.}
\label{deep_neural_network_illustration}
\end{figure}

Deep neural networks, often abbreviated as simply "networks", are a machine learning model which generally consists of several concatenated layers. The network processes the input data by propagating it through its layers for the purpose of performing a certain task.
When a network consists of many layers it is commonly referred to as a "deep neural network".
A conventional feed-forward neural network, which is the focus of the works we survey hereafter, has the following structure. It consists of $L$ layers, where the first $L-1$ layers are referred to as "hidden layers" and the $L$-th layer represents the network's output. Each layer in the network consists of several neurons (nodes).
An illustration of a neural network is given in Fig.~\ref{deep_neural_network_illustration}.

Feed-forward neural networks are networks in which the data propagates in a single direction, as opposed to other neural network models such as Recurrent Neural Networks (RNNs) in which the network connections form internal cycles.
Though many different variations of feed-forward neural networks exist, classically they either have fully connected layers or convolutional layers. A feed-forward neural network with at least one convolutional layer, in which at least one convolution kernel is used, is referred to as a Convolutional Neural Network (CNN).
In standard fully connected networks, every neuron in every layer is connected to each neuron in the previous layer. Such a connection is mathematically defined as a linear transformation using a weight matrix followed by the addition of a bias term, which is then followed by a non-linear activation function. The most commonly used activation functions are the rectified linear unit (ReLU), the sigmoid function and the hyperbolic tangent (tanh) function. These activation functions are described in Table~\ref{Activation_Functions}. There are other non-linearities that can be applied to a layer's output, for example pooling which decreases the layer's dimensions by aggregating information.

\begin{table}[h!]
\centering
\caption{Non-linear activation functions}
\label{Activation_Functions}
\begin{tabular}{|c|c|c|c|}
\hline
Name & Function: $\phi(x)$ & Derivative: $\frac{d\phi(x)}{dx}$ & Function output range \\
\hline\hline
ReLU & max\{0,x\}  & 1 if $x>0$; 0 if $x \leq 0$ & [0, $\infty$) \\
\hline
Sigmoid & $\frac{1}{1+e^{-x}}$ & $\phi(x)(1-\phi(x)) = \frac{e^{-x}}{(1+e^{-x})^2}$ & (0,1)\\
\hline
Hyperbolic tangent & $\frac{e^{x}-e^{-x}}{e^{x}+e^{-x}}$  & $1-\phi(x)^2 = 1 - \left( \frac{e^{x}-e^{-x}}{e^{x}+e^{-x}} \right)^2$ & (-1,1)\\
\hline
\end{tabular}
\end{table}

We use the index $l=1,\dots,L$ to denote a specific layer in the network, and $h_l$ to denote the amount of neurons in the $l$-th layer of the network. Accordingly $h_0 = D$ and $h_L = K$ represent the network's input and output dimensions respectively. The $h_l$-dimensional output of the $l$-th layer in the network is represented by the vector $z_{l}$.
The output of the last layer, $z_L$ is also denoted by $f \triangleq z_L$, representing the network's output.
The weight matrix of the $l$-th layer of a network is denoted by $W_{l} \in \mathbb{R}^{h_{l} \times h_{l-1}}$. We denote the element in the $i$-th row and the $j$-th column in the weight matrix of the $l$-th layer in the network by $w_{ij}^{l}$. The bias vector of the $l$-th layer is denoted by $b_{l} \in \mathbb{R}^{h_l}$. In addition, $vec \left( \{ W_l \}_{l=1}^L \right)$ is the column-stack vector representation of all of the network's weights. The activation function applied to every neuron in this layer is denoted by $\phi_{l}$, which is applied element-wise when its input is a vector. Consequently, the relation between two consecutive layers in a fully connected DNN is given by
\begin{align}
z_{l} = \phi_{l} \left( W_{l} z_{l-1} + b_{l}\right).
\end{align}
In most cases, the same non-linear activation function is chosen for all the layers of the network.

In classification tasks, a neural network can be used to classify an input to one of $K$ discrete classes, denoted using the index $k=1,\dots,K$. A common choice for the last layer of a network performing a classification task is the softmax layer, which transforms the output range to be between 0 and 1:
\begin{align}
f[k] = z_{L}[k] = \softmax \{ z_{L-1}[k] \} = \frac{e^{z_{L-1}[k]}}{\sum_{j=1}^{h_{L-1}} e^{z_{L-1}[j]}}.
\end{align}
The predicted class $k^*$ for an input $x$ is determined by the index of the maximal value in the output function obtained for this input sample, i.e.
\begin{align}
k^{*} = \argmax_{k} f[k].
\end{align}
Since for any $k=1,\dots,K$ the output range is $f[k] \in (0,1)$, the elements of $f \in \mathbb{R}^K $ (the network's output) are usually interpreted as probabilities assigned by the network to the corresponding class labels, and accordingly the class with the highest probability is chosen as the predicted class for a specific input. The usage of the softmax layer is usually coupled with the cross-entropy loss function defined by
\begin{align}
\ell(y,f(x)) = -\sum_{k=1}^K y[k]\log \left( f(x)[k] \right) = -\log \left( f(x)[k^*] \right),
\end{align}
where $k^*$ is the index of the correct class of the input $x$.

The training of a neural network involves the solution of an optimization problem which encapsulates the discrepancy between the true labels and the estimated labels of the training dataset, computed using a loss function.

Typically used loss functions are the cross-entropy loss defined above, the squared error loss (using the $\ell_2$ norm) defined by
\begin{align}
\ell(y,f(x)) = || y-f(x) ||_2^2 = \sum_{k=1}^K \left( y[k] - f(x)[k] \right)^2,
\end{align}
the absolute error loss (using the $\ell_1$ norm) defined by
\begin{align}
\ell(y,f(x)) = || y-f(x) ||_1 = \sum_{k=1}^K \left| y[k] - f(x)[k] \right|,
\end{align}
etc. In many cases, a closed form solution is unobtainable, whereas in other cases it is too computationally demanding. Therefore, the optimization problem is usually solved using some variant of the gradient descent algorithm, which is an iterative algorithm. In every iteration, a step is taken in the direction of the negative gradient vector of the loss function, i.e. the direction of the steepest descent, and the model parameters are updated accordingly. The size of the taken step is tuned using a scalar hyper-parameter, most commonly referred to as the "learning rate". The update of the model parameters in every iteration is given by
\begin{align}
\label{BGD}
W_n = W_{n-1} - \alpha \nabla_{W} \ell_{emp}(f,\mathbf{s}),
\end{align}
where $n$ represents the training iteration and $\alpha$ represents the learning rate.
In most cases, in every iteration only a subset of the training dataset is used to compute the gradient of the loss function. This subset is commonly referred to as a training "mini-batch".
Note that when the entire training dataset is used to evaluate the gradient of the loss, the optimization algorithm is usually referred to as "Batch Gradient Descent", whereas when a subset of the training dataset is used the optimization algorithm is referred to as "Stochastic Gradient Descent" (SGD).
The usage of training mini-batches is computationally beneficial as it requires the usage of less data in each training iteration. It has other advantages as well, as will be detailed later in this article.

The computation of the gradient of the loss function with respect to the model parameters, which is necessary to perform an optimization step, is typically a costly operation as the relation between the input and the output of a DNN is quite complex and cannot be expressed in a closed-form expression.
In most implementations this computation is based on the "back-propagation" algorithm, which is based on the chain rule for the derivation of the composition of functions \cite{Goodfellow16Deep}. This algorithm essentially computes the product of the partial derivatives along the different layers of the networks, propagating from the network's output to its input.

When training a neural network, and other machine learning algorithms as well, it is common practice to incorporate the usage of regularization techniques. This is effectively equivalent to making a prior assumption on the model itself or its input data.
Using regularization techniques introduces several benefits. Firstly, these techniques discourage the learned algorithm from over-fitting the training data, i.e. they encourage the learning algorithm to learn the underlying rule of the training data rather than memorize the specifics of the training dataset itself. This purpose is achieved by essentially penalizing the learned algorithm for being too complex and "artificially" fitting the specifics of the training data.
Secondly, regularization techniques promote the stability of the learned algorithm, in the sense that a small change to the input would not incur a large change to the output \cite{Understanding_Machine_Learning}.

Regularization techniques can generally be categorized as either explicit or implicit.
Explicit regularization techniques traditionally incorporate an additional loss function into the training objective function directly aimed at adapting the model to the assumed prior. When an additional regularization loss is used, the balance between the regularization term and the objective loss function is tuned using a multiplicative scalar hyper-parameter. Commonly used explicit regularization techniques include the incorporation of an additional loss in the form of the norm of the model weights (parameters), the usage of dropout, in which any neuron in the network is zeroed during training with a certain probability, and more.
Implicit regularization techniques have a more indirect influence on the learned function, and include a variety of techniques such as early stopping, data augmentation, and also model choices such as the used network architecture and optimization algorithm \cite{Goodfellow16Deep}.

\section{Generalization error in deep learning}
\label{Generalization Error in Deep Learning}

An important open question in machine learning research is what is the source of the generalization capabilities of deep neural networks. A better understanding of what affects this generalization error is essential towards obtaining reliable and robust deep neural network architectures with good performance.
Throughout the following subsections we review different theories, insights and empirical results that shed light on this topic. Several lines of work aim at bounding the generalization error of neural networks, whereas others seek for complexity measures that correlate to this error and explain what affects it. In addition, we review both works that characterize different aspects of the training phase, ranging from the size of the training mini-batch to the number of training iterations, and works that characterize the solution of the training optimization problem. These  works represent the most prominent lines of research in this field.

\subsection{Understanding deep learning requires rethinking generalization}
\label{Understanding deep learning requires rethinking generalization}

As described in section~\ref{Deep neural networks}, regularization techniques such as weight decay and dropout have been shown to improve the generalization capabilities of machine learning algorithms and specifically deep neural networks by preventing over-fitting to the training dataset.
Data over-fitting is a common problem when training deep neural networks since they are highly over-parameterized models which are usually trained using a small amount of data compared to the number of parameters in the model.
Regularization helps to reduce the model's complexity and thereby achieve a lower generalization error. For this reason, using regularization techniques is a common practice in the training of machine learning models.

In \cite{RethinkingGeneralization} some insight is given into the role of explicit and implicit regularization in reducing a network's generalization error. 
Different explicit regularization methods (such as data augmentation, weight decay and dropout) are empirically compared and the conclusion that follows is that explicit regularization is neither a sufficient nor a necessary technique to control the generalization error of a deep neural network. Namely, not using regularization during training does not necessarily mean a larger generalization error will be obtained.

As previously mentioned in section~\ref{The learning problem}, the Rademacher complexity is a complexity measure of a hypothesis set $\mathcal{H}$.
It is shown that adding explicit regularization to the training phase (e.g. dropout, weight decay) effectively reduces the Rademacher complexity of the hypothesis space of the possible solutions by confining it to a subspace of the original hypothesis space which has lower complexity. However, this does not necessarily imply a better generalization error, as in most cases the Rademacher complexity measure is not powerful enough to capture the abilities of deep neural networks.

Similarly, implicit regularization techniques, such as the usage of the SGD algorithm for training, early stopping and batch normalization \cite{Batch_Normalization}, may also play an important role in improving the generalization capabilities of deep neural networks. Yet, empirical findings show that they are not indispensable for obtaining good generalization. 

Deep neural networks can achieve a zero training error even when trained on a random labeling of the training data, meaning deep neural networks can easily fit random labels, which is indicative of very high model capacity. Expanding the scope of this premise, the relation to convolutional neural networks (CNNs) is made by showing that state-of-the-art CNNs for image classification can easily fit a random labeling of the training data, giving further support to the notion that deep neural networks are powerful enough to \emph{memorize} the training data, since randomly labeled data does not encapsulate any actual underlying rule.

These empirical findings are explained using a theoretical result, which shows that a two layer neural network already has perfect expressivity when the number of parameters exceeds the number of data samples. Specifically, there exists a two-layer neural network with ReLU activations and $2N+D$ weights that can represent any function on a sample of size $N$ in $D$ dimensions.

It follows that training remains a relatively easy task, even for random labels and for data that has been subject to different kinds of random shuffling and permutations, i.e. training is easy even when the model does not generalize well.
In addition, it has been empirically established that training on random labels only increases the training time by a small constant factor.

In this context it is important to note that extensive research efforts are still aimed at classical complexity measures such as the Rademacher complexity.
For example, \cite{size_independent_sample_complexity} provides a bound for the Rademacher complexity of DNNs assuming norm constraints on their weight matrices. Under several assumptions this bound is independent of the network size (width and depth).
According to this bound the Rademacher complexity is upper bounded by
\footnote{$\mathcal{\tilde{O}}$ is the upper bound to the complexity up to a logarithmic factor of the same term.}
\begin{align}
\mathcal{\tilde{O}} \left( R \sqrt{\frac{\log \{\frac{R}{\Gamma} \}}{\sqrt{N}}} \right),
\end{align}
where $R$ is an upper bound on the product of the Frobenius norms of the network's weight matrices and 
$\Gamma$ is a lower bound on the product of the spectral norms of the network's weight matrices.

The work in \cite{RethinkingGeneralization} generally emphasizes the need for a different approach in examining the generalization of DNNs. One specific incentive is the refuted common notion that the widely used regularization techniques are a necessary condition for obtaining good generalization. Moreover, even though several measures have been shown to be correlated with the generalization error of DNNs (as shown in \cite{ExploringGeneralization} and discussed in subsection~\ref{Exploring Generalization in Deep Learning}), there is still a need for tighter bounds and more explanations for the generalization capabilities of DNNs. Therefore, a more comprehensive theory is necessary to explain why DNNs generalize well even though they are capable of memorizing the training data.

\subsection{Exploring generalization in deep learning}
\label{Exploring Generalization in Deep Learning}

In \cite{ExploringGeneralization} several different measures and explanations for the generalization capabilities of DNNs are examined.
The examined measures include norm-based control, robustness and sharpness, for which a connection to the PAC-Bayes theory is drawn.
The different measures are evaluated based on their theoretical ability to guarantee generalization and their performance when empirically tested.

{\em The capacity of a model} for several given metrics (e.g., $\ell_2$ distance), which is examined throughout the work in \cite{ExploringGeneralization}, represents the number of training examples necessary to ensure generalization, meaning that the test error is close to the training error.

With similarity to commonly established notions on the matter, it is claimed that using a VC-dimension measure to provide a bound on the capacity of neural networks is insufficient to explain their generalization abilities.
Relying on the works in \cite{Nearly_tight_VC_dimension} and \cite{AlmostLinear}, a bound is proposed on the VC-dimension of feed-forward neural networks with ReLU activations in terms of the number of parameters in the network. This bound is given by
\begin{align}
\mathrm{VCdim} = \tilde{\mathcal{O}} (L \cdot \mathrm{dim}(W)),
\end{align}
where $\mathrm{dim}(W)$ is the number of parameters in the network and $L$ is the amount of layers in the network. This bound is very loose and therefore fails to explain the generalization behavior of these networks. Neural networks are highly over-parameterized models, but they can fit a desired function almost perfectly with a training set much smaller than the amount of parameters in the model. For this reason, a bound which is linear in the amount of parameters in the model is too weak to explain the generalization behavior of deep neural networks.
We refer the reader to \cite{Anthony_Bartlett_NN_Learning} for some earlier work on the VC-dimension of neural networks.

Norm based complexity measures for neural networks with ReLU activations are presented as well. These measures do not explicitly depend on the amount of parameters in the model and therefore have a better potential to represent its capacity.

Relying on the work in \cite{Norm-BasedCapacityControl} four different norm based measures are used to evaluate the capacity of deep neural networks. These measures use the following lenient version of the definition of a classification margin: $\gamma_{margin}$ is the lowest value of $\gamma$ such that $\lceil{\epsilon N}\rceil$ data samples have a margin lower than $\gamma$, where $\epsilon > 0$ is some small value and $N$ is the size of the training set. The four measures are given in Table~\ref{capacity_metrics_tables}.

\begin{table}
\caption{Norm-based capacity measures for the analysis of the generalization of DNNs}
\label{capacity_metrics_tables}
\centering
\begin{tabular}{|c|c|}
\hline
Capacity type & Capacity order \\[0.5ex]
\hline\hline
$\ell_2$-norm & $\frac{1}{\gamma_{margin}^2} \prod_{l=1}^{L} 4 ||W_l||_F^2$ \\
\hline
$\ell_1$ path-norm & $\frac{1}{\gamma_{margin}^2} \left(\sum_{j \in \prod_{k=0}^{L} [h_k]} \left| \prod_{l=1}^{L} 2W_l[j_l,j_{l-1}] \right| \right)^2$ \\
\hline
$\ell_2$ path-norm & $\frac{1}{\gamma_{margin}^2} \sum_{j \in \prod_{k=0}^{L} [h_k]} \prod_{l=1}^{L} 4h_l W_l^2[j_l,j_{l-1}]$\\
\hline
Spectral-norm & $\frac{1}{\gamma_{margin}^2} \prod_{l=1}^{L} h_l ||W_l||_2^2$ \\
\hline
\end{tabular}
\end{table}

$[h_k]$ is the set $\{h_1,...,h_k\}$ and $\prod_{k=0}^L[h_k]$ is the Cartesian product over the sets $[h_k]$. We remind the reader that $h_k$ is the amount of neurons in the k-th layer of the network, and accordingly $h_0=D$ where $D$ is the network's input dimension and $h_L=K$ where $K$ is the network's output dimension. Note that the $\ell_1$ and $\ell_2$ path-norms (see Table~\ref{capacity_metrics_tables}) sum over all possible paths going from the input to the output of the network, passing through one single neuron in each layer. Accordingly, the index $j$ represents a certain path consisting of one neuron in each layer, $j_l$ denotes the neuron in the $l$-th layer in the $j$-th path and $W_l[j_l,j_{l-1}]$ is the weight parameter (scalar) relating the $j_{l-1}$ and the $j_{l}$ neurons.

The results of empirical tests are presented in \cite{ExploringGeneralization}. These show that the training error is zero and all the aforementioned norm measures are larger for networks trained to fit random labels than for networks trained to fit the true labels of the data (specifically, the VGG network is used along with the CIFAR-10 dataset). These findings indicate that these norm-based measures can explain the generalization of DNNs, as the complexity of models trained on random labels is always higher than the complexity of models trained on the true labels, corresponding to the favorable generalization abilities of the latter.

Another capacity measure which is examined is the Lipschitz constant of a network. The question whether controlling the Lipschitz constant of a network with respect to its input can lead to controlling its capacity is investigated, relying on the relation between the weights of a network and its Lipschitz constant.
Different bounds on models' complexity based on their Lipschitz constants are reviewed. The relation between a network's Lipschitz constant and the norm of its weights is made, and it is shown that all the bounds relying on the Lipschitz constant result in very loose capacity bounds which are exponential in both the input dimension and the depth of the network.
The conclusion is that simply bounding the Lipschitz constant of a network is not enough to obtain a reasonable capacity control for a neural network, and that these bounds are merely a direct consequence of bounding the norm of the network's weights. 

In addition, another capacity measure which has been linked to the generalization abilities of DNNs is addressed: the sharpness of the obtained minimizer to the optimization problem of the training of DNNs. It is claimed that the notion of sharpness, which was formulated in \cite{Large_Batch_Training} (see Definition~\ref{Sharpness_definition}), cannot by itself capture the generalization behavior of neural networks. Instead, a related notion of \emph{expected sharpness} in the context of the PAC-Bayes theorem, combined with the norm of the network's weights, does yield a capacity control that offers good explanations to the observed phenomena.

In a recent related paper \cite{Over_Parametrization_and_Generalization}, a new complexity measure based on unit-wise capacities is proposed. A unit's capacity is defined by the $\ell_2$ norm of the difference between the input weights associated with this unit and the initialization values of these weights. These input weights are the multiplicative weights which were applied to the values of the units of the previous layer in order to obtain the value of this unit (i.e. the row that corresponds to this specific unit in the weight matrix of this unit's layer). This complexity measure results in a tighter generalization bound for two layer ReLU networks.

In another recent work \cite{HighDimensionalApproximationEstimation} a bound for the statistical error (test error) is given for networks with $\ell_1$ constraints on their weights and with ReLU activations. In this setting the test error is shown to be upper bounded by $\sqrt{\frac{L^3 \log D}{N}}$, where in this context $D$ is the maximal input dimension to a layer in the network.
With this bound the input dimension can be much larger than the amount of samples, and the learned algorithm would still be accurate if the target function has the appropriate $\ell_1$ constraints and $N$ is large compared to $L^3 \log d$.

Following the above, it can be concluded that even though various measures of the capacity of DNNs exist in the literature, some exhibiting good correlation with the generalization abilities of DNNs, a comprehensive theoretical formulation with adequate empirical results that explain the generalization abilities of DNNs is still an active field of research.

\subsection{A PAC-Bayesian approach to spectrally-normalized margin bounds for neural networks}
\label{A PAC-Bayesian approach to spectrally-normalized margin bounds for neural networks}

In a more recent work \cite{PacBayesianApproachMarginBounds} a generalization bound for neural networks with ReLU activations is presented in terms of the product of the spectral norm and the Frobenius norm of their weights.
A bound on the changes in a network's output with respect to a perturbation in its weights is used to derive a generalization bound.
The perturbation bound relies on the following constraint on the input domain: 
\begin{eqnarray}
\mathcal{X}_{B,D} = \left\{ x \in \mathbb{R}^{D} | \sum_{i=1}^{D} x_{i}^2 \leq B^2  \right\}, 
\end{eqnarray}
and is formulated in the following lemma.
\begin{lem}(Perturbation Bound)
\label{Lemma_perturbation_bound}
For any $B, L > 0$, let $f_{W} : \mathcal{X}_{B,D} \rightarrow \mathbb{R}^{K}$ be an $L$-layer neural network with ReLU activations. Then for any $W = vec (\{ W_l \}_{l=1}^L)$, and $x \in \mathcal{X}_{B,D}$, and any perturbation $U = vec ({ \{ U_l \}}_{l=1}^{L} )$ such that $|| U_l ||_2 \leq \frac{1}{L} ||W_l||_2$, the change in the output of the network can be bounded as follows:
\begin{align}
||f_{W+U}(x) - f_{W}(x)||_2 \leq eB \left( \prod_{l=1}^{L} ||W_{l}||_2 \right) \sum_{l=1}^L \frac{||U_l||_2}{||W_l||_2}.
\end{align}
\end{lem}

This bound makes a direct relation between a perturbation in the model parameters and its effect on the network's output. Therefore, it leads to an upper bound on the allowed perturbation in the model parameters for a desired margin $\gamma$. The consequent generalization bound for neural networks with ReLU activations is given in the following theorem.
\begin{thm}
\label{thm_margin_bound_Relu}
For any $B,L,\alpha > 0$, let $f_{W} : \mathcal{X}_{B,D} \rightarrow \mathbb{R}^{K}$ be an $L$-layer feed-forward network with ReLU activations. Then, for any constant $\delta$, margin $\gamma >0$, network parameters $W = vec (\{ W_l \}_{l=1}^L)$ and training set of size $N$, we have with probability exceeding $1-\delta$ that
\begin{align}
\label{margin_bound}
\ell_{exp,0}(f_{W}) \leq \ell_{emp,\gamma}(f_W)+O \left( \sqrt{\frac{B^2L^2 \alpha \ln (L \alpha) \prod_{l=1}^{L} ||W_l||_2^2 \sum_{l=1}^{L} \frac{||W_l||_F^2}{||W_l||_2^2} +\ln{\frac{LN}{\delta}} }{\gamma^2 N}} \right),
\end{align}
where $\alpha$ is an upper bound on the number of units in each layer in the network.
\end{thm}
The proof of this theorem relies on the PAC-Bayes theory. We refer the reader to \cite{PacBayesianApproachMarginBounds} for the full proofs of both Lemma~\ref{Lemma_perturbation_bound} and Theorem~\ref{thm_margin_bound_Relu}.

As would have been expected, the bound in \eqref{margin_bound} becomes tighter as the margin $\gamma$ increases. In addition, the bound is looser as the number of layers in the network $L$ increases
and for a domain with a larger parameter $B$, as it encapsulates a larger input domain. With similarity to other generalization error bounds, this bound becomes tighter as the size of the training set increases.

This bound takes a step further towards a comprehensive explanation to the generalization capabilities of DNNs, as unlike many other bounds it incorporates several different measures that define a DNN: the number of layers in the network, the spectral norm of the layers' weights, the number of neurons in each layer, the size of the input domain, the required classification margin and the size of the training set.

Another prominent work \cite{Spectrally-normalized_margin_bounds} provides a margin-based bound for the generalization error of DNNs which scales with their spectral complexity, i.e. their Lipschitz constant (the product of the spectral norms of the weight matrices) times a correction factor. This bound relies on the following definition for the spectral complexity of a network.
\begin{defn} (Spectral Complexity)
A DNN with reference matrices $(M_1,\dots,M_L)$ with the same dimensions as the weight matrices $W_1,\dots,W_L$, for which the non-linearities $\phi_i$ are $\rho_i$-Lipschitz respectively, has spectral complexity
\begin{align}
R_{W} = \left( \prod_{i=1}^{L} \rho_i ||W_i||_{2}  \right)
\left( \sum_{i=1}^{L} \frac{||W_i^T - M_i^T||_{2,1}^{\frac{2}{3}}}{||W_i||_2^{\frac{2}{3}}} \right)^{\frac{3}{2}},
\end{align}
where here $||\cdot||_2$ represents the spectral norm and the $(p,q)$-matrix norm is defined by $||\cdot||_{p,q} = || (||A_{:,1}||_p, \dots, ||A_{:,m}||_p)||_q$ for some matrix $A \in \mathbb{R}^{d \times m}$.
\end{defn}
Note that the spectral complexity depends on the chosen reference matrices. This leads to the following margin-based generalization bound. For a DNN with training set $\{(x_i,y_i)\}_{i=1}^N$ drawn i.i.d. from some distribution $\mathcal{D}$, with weight matrices $W_1,\dots,W_L$, for every margin $\gamma > 0$, with probability at least $1-\delta$ it holds that
\begin{align}
\ell_{exp,0}(f) \leq
\ell_{emp,\gamma} (f,x) +
\mathcal{\tilde{O}} \left( \frac{||X||_2 R_W}{\gamma N} \log (\max_i h_i) + \sqrt{\frac{\log (1 / \delta)}{N}} \right)
\end{align}
where $||X||_2 = \sqrt{\sum_{i} ||x_i||_2^2}$, and $\max_i h_i$ represents the maximal amount of neurons in any layer in the DNN. The proof is left to the paper \cite{Spectrally-normalized_margin_bounds}.

The PAC-Bayes framework was also examined in \cite{ComputingNonvacuous}, where a PAC-Bayes bound on a network's generalization error is optimized in order to obtain non-vacuous (tight, non-trivial) generalization bounds for deep stochastic neural network classifiers.
It is hypothesized that the SGD optimization algorithm obtains good solutions only if these solutions are surrounded by a relatively large volume of additional solutions that are similarly good. This leads to the notion that the PAC-Bayes bound has the potential to provide non-vacuous bounds on the generalization error of deep neural networks when it is used optimize the stochastic classifier.  
As shown in section~\ref{The learning problem}, the PAC-Bayes theorem bounds the expected loss of a stochastic classifier using the Kullback-Leibler divergence between an a-priori probability distribution $P$ and a posterior probability distribution $Q$, from which a classifier is chosen when the training data is available.

An optimization is performed over the distributions $Q$, in order to find the distribution that minimizes the PAC-Bayes bound. This is done using a variation of the stochastic gradient descent algorithm and using a multivariate Gaussian posterior $Q$ on the network parameters. In each step the network's weights and their corresponding variances are updated by a step in the direction of an unbiased estimate of the gradient of an upper bound on the PAC-Bayes bound. Using this approach, a finer bound on the generalization error of neural networks is obtained. 
We refer the reader to \cite{ComputingNonvacuous} for the exact mathematical formulation.
This approach relates to similar notions examined in other works (such as \cite{Large_Batch_Training,SharpMinima}), that make a connection between the sharpness of the solution obtained using the SGD algorithm and its ability to generalize well.

\subsection{Stability and generalization}
\label{Stability and Generalization}

The \emph{stability} of a learning algorithm is an important characteristic which  represents its ability to maintain similar generalization results when a training example is excluded or replaced in the training dataset.
In \cite{Stability_and_Generalization} a sensitivity-driven approach is used to derive generalization error bounds. The sensitivity of learning algorithms to changes in the training set, which are caused by sampling randomness and by noise in the sampled measurements, is formally defined and analyzed throughout \cite{Stability_and_Generalization}.

A stable learning algorithm is an algorithm which is \emph{not} sensitive to small changes in the training set, i.e. an algorithm for which a small change in its training set results in a small change in its output. As mentioned earlier, such a small change can be the exclusion or replacement of a certain training example.

Statistical tools of concentration inequalities lead to bounds on the generalization error of stable learning algorithms when the generalization error is essentially treated as a random variable whose expected value is zero when it is constrained to be roughly constant.

The following are several useful definitions for the examination of the influence of changes in the training set $\mathbf{s}$ which consists of $N$ samples independently drawn from the distribution $\mathcal{D}$.
\begin{itemize}
\item
By excluding (removing) the $i^{th}$ sample in the training set the following set is obtained $\mathbf{s}^{\backslash i} = \{s_1,\dots,s_{i-1},s_{i+1},\dots,s_{N} \}$.
\item
By replacing the $i^{th}$ sample in the training set the following set is obtained $\mathbf{s}^{i} = \{s_1,\dots,s_{i-1},s_{i}',s_{i+1},\dots,s_{N} \}$, where the new sample $s_{i}'$ is drawn from the same distribution $\mathcal{D}$ and is independent from $\mathbf{s}$.
\end{itemize}

The following analysis is based on inequalities that relate moments of multi-dimensional random functions to their first order finite differences.
Let us present the following four definitions for the stability of a learning algorithm. These definitions will be later used to derive generalization bounds.
\begin{defn}
An algorithm $\mathcal{A}$ has hypothesis stability $\beta$ with respect to the loss function $\ell$ if the following holds
\begin{align}
\forall i \in \{1,\dots,N\}, \mathbb{E}_{\mathbf{s},s} \left[ | \ell ( \mathcal{A}_{\mathbf{s}},s) - \ell ( \mathcal{A}_{\mathbf{s^{\backslash i}}},s) | \right] \leq \beta. 
\end{align}
\end{defn}

The hypothesis stability relates to the average change caused by the exclusion of one training sample. In order to limit the change at every specific training point, the following definition of point-wise hypothesis stability is presented.
\begin{defn}
An algorithm $\mathcal{A}$ has point-wise hypothesis stability $\beta$ with respect to the loss function $\ell$ if the following holds
\begin{align}
\forall i \in \{1,\dots,N\}, \mathbb{E}_{\mathbf{s}} [ | \ell ( \mathcal{A}_{\mathbf{s}},s_i) - \ell ( \mathcal{A}_{\mathbf{s^{\backslash i}}},s_i) | ] \leq \beta. 
\end{align}
\end{defn}

For measuring the change in the expected error of an algorithm instead of the point-wise change, the following definition of error stability, which satisfies a weaker notion of stability, is presented. 
\begin{defn}
An algorithm $\mathcal{A}$ has error stability $\beta$ with respect to the loss function $\ell$ if the following holds
\begin{align}
\forall \mathbf{s} \in \mathcal{Z}^N, \forall i \in \{1,\dots,N\}, \left| \mathbb{E}_{s} [ \ell ( \mathcal{A}_{\mathbf{s}},s) ] - \mathbb{E}_{s} [ \ell ( \mathcal{A}_{\mathbf{s^{\backslash i}}},s) ] \right| \leq \beta. 
\end{align}
\end{defn}

Lastly, the uniform stability, which is a stronger definition of stability, leads to tight bounds.
\begin{defn}
An algorithm $\mathcal{A}$ has uniform stability $\beta$ with respect to the loss function $\ell$ if the following holds
\begin{align}
\forall \mathbf{s} \in \mathcal{Z}^N, \forall i \in \{1,\dots,N\}, || \ell( \mathcal{A}_{\mathbf{s}},s) -  \ell ( \mathcal{A}_{\mathbf{s^{\backslash i}}},s) ||_{\infty} \leq \beta. 
\end{align}
\end{defn}

Using these four definitions of the stability of learning algorithms, the following bounds on the relation between the empirical loss and the expected loss are derived. Let us denote the leave-one-out loss on the training set by $\ell_{loo}(\mathcal{A}_{\mathbf{s}}) \triangleq \frac{1}{N} \sum_{i=1}^{N} \ell \left( \mathcal{A}_{\mathbf{s}^{\backslash i}}, s_i \right)$.
This loss is of importance when discussing an algorithm's stability since it represents its average test loss on a specific sample when it is excluded from its training set.

The following are polynomial bounds on the expected loss.
\begin{thm}
For any learning algorithm $\mathcal{A}$ with hypothesis stability $\beta_{1}$ and point-wise hypothesis stability $\beta_2$ with respect to a loss function $\ell$ such that $0 \leq \ell(f(x),y) \leq M$, we have with probability $1-\delta$,
\begin{align}
\ell_{exp}(\mathcal{A}_{\mathbf{s}}) \leq \ell_{emp}(\mathcal{A}_{\mathbf{s}}) + \sqrt{\frac{M^2 + 12MN\beta_2}{2N\delta}},
\end{align}
and
\begin{align}
\ell_{exp}(\mathcal{A}_{\mathbf{s}}) \leq \ell_{loo}(\mathcal{A}_{\mathbf{s}}) + \sqrt{\frac{M^2 + 6MN\beta_1}{2N\delta}}.
\end{align}
\end{thm}

We refer the reader to \cite{Stability_and_Generalization} for the proofs.
Specifically, for the regression and classification cases, bounds based on the uniform stability of learning algorithms are derived. The bounds for the latter case are left to the original paper, whereas the bounds for the former case are as follows.
\begin{thm}
\label{uniform_stability_bounds}
Let $\mathcal{A}$ be an algorithm with uniform stability $\beta$ with respect to a loss function $\ell$ such that $0 \leq \ell(\mathcal{A_{\mathbf{s}}},s) \leq M$, for all $s \in \mathcal{Z}$ and all sets $\mathbf{s}$. Then, for any $N \geq 1$, and any $\delta \in (0,1)$, the following bounds hold (separately) with probability at least $1-\delta$ over the random draw of the sample $\mathbf{s}$,
\begin{align}
\ell_{exp}(\mathcal{A}_{\mathbf{s}}) \leq \ell_{emp}(\mathcal{A}_{\mathbf{s}}) + 2\beta + (4N\beta+M)\sqrt{\frac{\ln(1/\delta)}{2N}},
\end{align}
and
\begin{align}
\ell_{exp}(\mathcal{A}_{\mathbf{s}}) \leq \ell_{loo}(\mathcal{A}_{\mathbf{s}}) + \beta + (4N\beta+M)\sqrt{\frac{\ln(1/\delta)}{2N}}.
\end{align}
\end{thm}

This theorem gives tight bounds when the stability $\beta$ scales as $\frac{1}{N}$, which is the case for several prominent algorithms
such as the $K$-NN classifier with respect to the \{0,1\} loss function, and the SVM classifier with respect to the Hinge loss function.
Bounds for the case when regularization is used can be controlled by the regularization parameter (a scalar hyper-parameter, usually denoted by $\lambda$, which controls the weight of the regularization term in the objective loss function) and can therefore be very tight. These bounds are left to the original paper \cite{Stability_and_Generalization}.

The specific relation to deep neural networks was made in \cite{Train_Faster_Generalize_Better}, where several theorems regarding the stability of deep neural networks are given.
It is shown that stochastic gradient methods, which are the most commonly used methods for training DNNs, are stable. Specifically, the following theorem establishes that stochastic gradient methods are uniformly stable.
\begin{thm}
Assume that $\ell(x) \in [0,1]$ is an M-Lipschitz and $\epsilon-smooth$ loss function for every $x$. Suppose that we run the stochastic gradient method for $T$ steps with monotonically non-increasing step sizes $\alpha_{t} \leq \frac{c}{t}$. Then, the stochastic gradient method applied to $\ell$ has uniform stability with
\begin{align}
\beta_{stability} \leq \frac{1+\frac{1}{\epsilon c}}{N-1} (2cM^2)^{\frac{1}{\epsilon c + 1}} T^{\frac{\epsilon c}{\epsilon c + 1}},
\end{align}
where a function $\ell(x)$ is M-Lipschitz if for all points $x$ in the domain of $\ell$ it holds that $|| \nabla \ell (x) ||_2 \leq M$, and a function $\ell(x)$ is $\epsilon-smooth$ if for all $x, \hat{x}$ in the domain of $\ell$ it holds that $|| \nabla \ell(x) - \nabla \ell(\hat{x}) ||_2 \leq \epsilon ||x - \hat{x}||_2$.
\end{thm}

Specifically, if the constant factors that depend on $\epsilon$, $c$ and $M$ are omitted, the bound on the uniform-stability is given by
\begin{align}
\beta_{stability} \lessapprox \frac{1}{N} T^{1-\frac{1}{\epsilon c + 1}}.
\end{align}
This bound implies that under certain assumptions on the loss function the uniform stability $\beta$ scales as $\frac{1}{N}$ for deep neural networks, and in this case the same bounds from Theorem~\ref{uniform_stability_bounds} are tight for deep neural networks as well.
We refer the reader to \cite{Train_Faster_Generalize_Better} for the formal proof of this theorem.

The notion of stability has therefore been shown to be of importance in the evaluation of a learning algorithm's generalization error. It has been established that stable algorithms yield a lower expected loss and therefore a lower generalization error, particularly for deep neural networks which in many cases can obtain a training loss of zero (as shown in \cite{RethinkingGeneralization}).
A comprehensive overview of \cite{Train_Faster_Generalize_Better} is given in subsection~\ref{Train faster, generalize better: Stability of stochastic gradient descent}.

\subsection{Robustness and generalization}
\label{Robustness and Generalization}

In a later work \cite{RobustnessGeneralization} a notion from robust optimization theory is used to examine the generalization capabilities of learning algorithms with respect to their robustness. An algorithm is said to be robust if it achieves similar performance on a test sample and a training sample which are close in some sense, i.e. if a test sample is similar to a training sample, then its corresponding test error is close to the corresponding training error. This means that a robust learning algorithm is not sensitive to small perturbations in the training data. This notion applies to general learning algorithms, not only deep neural networks. The formal definition of algorithm robustness is given in section~\ref{The learning problem}.
The following is the generalization error bound for a robust learning algorithm.

\begin{thm}
If $\mathbf{s}$ consists of $N$ i.i.d. samples, and $\mathcal{A}$ is $(K, \epsilon(\mathbf{s}))$-robust, then for any $\delta > 0$, with probability at least $1-\delta$,
\begin{align}
|\ell_{exp}(\mathcal{A_{\mathbf{s}}})-\ell_{emp}(\mathcal{A_{\mathbf{s}}})| \leq \epsilon(\mathbf{s}) +
M \sqrt{\frac{2K\ln{2}+2\ln{(1/\delta)}}{N}}.
\end{align}
\end{thm}
This holds under the assumption that $\ell(\mathcal{A_\mathbf{s}})$ is non-negative and upper-bounded uniformly by the scalar $M$.

A new relaxed definition of pseudo-robustness (weak robustness) is proposed. Pseudo-robustness is both a necessary and sufficient condition for asymptotic generalizability of learning algorithms in the limit superior sense (as shown in Definition~\ref{Generalization_definition_robustness} and Definition~\ref{Generalization_definition_pseudo_robustness} hereafter).
Under the definition of pseudo-robustness, the condition for robustness as mentioned in \eqref{Robustness_Definition} in the preliminaries, only has to hold for a subset of the training samples.
The definition of pseudo-robustness is as follows.

\begin{defn}
Algorithm $\mathcal{A}$ is $(K,\epsilon(\mathbf{s}),\hat{N})$ pseudo-robust if $\mathcal{Z}$ can be partitioned into $K$ disjoint sets, denoted as $\{ C_i \}_{i=1}^{K}$, and a subset of training samples $\hat{\mathbf{s}}$ with $|\hat{\mathbf{s}}| = \hat{N}$ such that $\forall s \in \hat{\mathbf{s}}$,

\begin{align}
s, z \in C_i, \Rightarrow |\ell(\mathcal{A_{\mathbf{s}}},s)-\ell(\mathcal{A_{\mathbf{s}}},z)| \leq \epsilon(\mathbf{s}).
\end{align}
\end{defn}

The following theorem gives a bound on the generalization error of pseudo-robust learning algorithms.

\begin{thm}
If $\mathbf{s}$ consists of $N$ i.i.d. samples, and $\mathcal{A}$ is $(K,\epsilon(\mathbf{s}),\hat{N})$ pseudo-robust, then for any $\delta > 0$, we have that with a probability at least $1-\delta$,
\begin{align}
\left| \ell_{exp}(\mathcal{A_\mathbf{s}})-\ell_{emp}(\mathcal{A_\mathbf{s}}) \right|
\leq \frac{\hat{N}}{N} \epsilon(s) +
M \left(\frac{N-\hat{N}}{N} + \sqrt{\frac{2K\ln{2}+2\ln{(1/\delta)}}{N}} \right).
\end{align}
\end{thm}
This holds under the assumption that $\ell(\mathcal{A_\mathbf{s}})$ is non-negative and upper-bounded uniformly by a scalar $M$.
The proof is offered in its entirety in \cite{RobustnessGeneralization}.

Robustness is an essential property for successful learning. In particular, pseudo-robustness (weak robustness) is indicative of the generalization abilities of a learning algorithm. A learning algorithm generalizes well if and only if it is pseudo-robust. This conclusion is formalized by the following definitions.

\begin{defn}
\label{Generalization_definition_robustness}
1. A learning algorithm $\mathcal{A}$ generalizes w.r.t. $\mathbf{s}$ if
\begin{align}
\limsup_{N} \left\{ \mathbb{E}_{t} \left( \ell(\mathcal{A}_{\mathbf{s}_{N}},t) \right) - \frac{1}{N} \sum_{i=1}^{N} \ell(\mathcal{A}_{\mathbf{s}_{N}},s_i) \right\} \leq 0.
\end{align}
2. A learning algorithm $\mathcal{A}$ generalizes with probability $1$ if it generalizes w.r.t. almost every $\mathbf{s}$.
\end{defn}

\begin{defn}
\label{Generalization_definition_pseudo_robustness}
1. A learning algorithm $\mathcal{A}$ is weakly robust w.r.t. $\mathbf{s}$ if there exists a sequence of $\{ \mathcal{D}_{N} \subseteq \mathcal{Z}^{N} \}$ such that $Pr\left( \mathbf{t}_{N} \in \mathcal{D}_{N} \right) \rightarrow 1$, and
\begin{align}
\label{weak_robustness_conclusion}
\limsup_{N} \left\{ \max_{\hat{\mathbf{s}}_{N} \in \mathcal{D}_N} \left[ \frac{1}{N} \sum_{i=1}^{N} \ell(\mathcal{A}_{\mathbf{s}_{N}},\hat{s_i})- \frac{1}{N} \sum_{i=1}^{N} \ell(\mathcal{A}_{\mathbf{s}_{N}},s_i) \right] \right\} \leq 0.
\end{align}
2. A learning algorithm $\mathcal{A}$ is asymptotically weakly robust if it is robust w.r.t. almost every $\mathbf{s}$.
\end{defn}
Note that $\mathcal{A}_{\mathbf{s}_N}$ is the learning algorithm $\mathcal{A}$ trained on the set $\mathbf{s}_N = \left\{ {s_1,\dots,s_N} \right\}$, and $\mathbf{\hat{s}}_N = \{ \hat{s}_1, \dots,
\hat{s}_N \} \in \mathcal{D}_N$ is the sequence of samples.
It follows from \eqref{weak_robustness_conclusion} that if for a large subset of $\mathcal{Z}^N$ the test error is close to the training error, then the learning algorithm is pseudo-robust (weakly robust). The thorough proof is offered in \cite{RobustnessGeneralization}.

The following theorem is given to make a general relation between pseudo-robustness of a learning algorithm and its generalization capabilities.
\begin{thm}
An algorithm $\mathcal{A}$ generalizes w.r.t. $\mathbf{s}$ if and only if it is weakly robust w.r.t. $\mathbf{s}$.
\end{thm}

The following corollary stems from the aforementioned theorem and further formalizes the discussed relation. 
\begin{cor}
An algorithm $\mathcal{A}$ generalizes with probability 1 if and only if it is asymptotically weakly robust. 
\end{cor}
Therefore, it has been established that weak robustness is a fundamental characteristic for learning algorithms to be able to generalize well.

In order to make a relation between the above theorems and feed-forward neural networks we introduce the \emph{covering number} term as defined in \cite{CoveringNumberDefinition}. 

\begin{defn} (Covering number)
For a metric space $\mathcal{S}$ with metric $d$ and $\mathcal{X} \subset \mathcal{S}$, it is said that $\hat{\mathcal{X}} \subset \mathcal{S}$ is a $\rho$-cover of $\mathcal{X}$ if $\forall x \in \mathcal{X}, \exists \hat{x} \in \hat{\mathcal{X}}$ such that $d(x,\hat{x}) \leq \rho$. The $\rho$-covering number of the space $\mathcal{X}$ with $d$-metric balls of radius $\rho$ is
\begin{align}
\mathcal{N}(\rho, \mathcal{X}, d) = \min\{|\hat{\mathcal{X}}| \text{ s.t. $\hat{\mathcal{X}}$ is a $\rho-$cover of $\mathcal{X}$\}}.
\end{align}
\end{defn}
Accordingly, the term $\mathcal{N}(\frac{\gamma}{2},\mathcal{Z},||\cdot||_{\infty})$, which is used in the following example, represents the $\frac{\gamma}{2}$-covering number of the space $\mathcal{Z}$ with the metric $||\cdot||_{\infty}$.

The following example makes the relation to deep neural networks.
\begin{exmpl}
Let $\mathcal{Z}$ be compact and the loss function on the sample $s=(x,y)$ be $\ell(\mathcal{A}_{\mathbf{s}},s) = \left| y-\mathcal{A}_{\mathbf{s}}(x) \right|$. Consider the $L$-layer neural network trained on $\mathbf{s}$, which is the following predicting rule given an input $x \in \mathcal{X}$
\begin{align}
\label{dnn_robustness_1}
\forall l=1,\dots,L-1:  x_i^l \triangleq \phi \left( \sum_{j=1}^{h_{l-1}} w_{ij}^{l-1} x_{j}^{l-1} \right); i=1,\dots,h_l;
\end{align}
\begin{align}
\label{dnn_robustness_2}
\mathcal{A}_{\mathbf{s}}(x) \triangleq \phi \left( \sum_{j=1}^{h_{L-1}} w_{j}^{L-1} x_{j}^{L-1} \right).
\end{align}
If there exist $\alpha$, $\beta$ such that the $L$-layer neural network satisfies that $|\phi(a)-\phi(b)| \leq \beta|a-b|$, and $\sum_{j=1}^{h_l} |w_{ij}^{l}|\leq \alpha$ for all $l,i$, then it is $\left( \mathcal{N}(\frac{\gamma}{2},\mathcal{Z},||\cdot||_{\infty}),\alpha^{L}\beta^{L}\gamma \right)$-robust, for all $\gamma > 0$.
\end{exmpl}
Note that $x^0 \triangleq x$ represents the network's input, and equations~\eqref{dnn_robustness_1}, \eqref{dnn_robustness_2} depict standard data propagation through the network.
An interesting result is that the number of neurons in each layer does not affect the robustness of the algorithm, and as a result the test error.

In \cite{Sokolic17Robust} the notion of robustness from \cite{RobustnessGeneralization} is used to derive a bound on the generalization error of DNN classifiers trained with the 0-1 loss, where the sample space $\mathcal{X}$ is a subset of a $C_M$ regular $D$-dimensional manifold whose covering number is upper bounded by $\mathcal{N} (\rho,\mathcal{X},d) \leq (\frac{C_M}{\rho})^D$.
In this case, the advantage of the robustness framework is that it provides a connection between the generalization error of the classifier and the data model. Yet, the bound provided in \cite{Sokolic17Robust} scales exponentially with the intrinsic dimension of the data. Therefore, it is a rather loose bound and a tighter bound is required to better explain the generalization capabilities of DNN classifiers.

\subsection{Stronger generalization bounds for deep nets via a compression approach}
\label{Stronger generalization bounds for deep nets via a compression approach}

A compression-based approach has recently been proposed to derive tight generalization bounds for deep neural networks \cite{StrongerBoundsCompression}.
This proposed compression is essentially a re-parameterization of the trained neural network, which relies on compression algorithms for reducing the effective number of parameters in deep neural networks.
Using noise stability properties a theoretical analysis of this compression-based approach leads to tight generalization bounds. Generalization bounds that apply to convolutional neural networks (CNNs) are also drawn for these compressed networks, and the correlation to their generalization capabilities is empirically established.

Firstly, it is shown in \cite{StrongerBoundsCompression} that Gaussian noise injected to different layers in a neural network has a rapidly decaying impact on the following layers. This attenuation of the noise as it propagates through the network layers implies a noise stability that allows the compression of individual layers of the network.
The definition of a compressible classifier is given in section~\ref{The learning problem}.

Incorporating the use of a "helper string" $s$,
which is essentially a vector of fixed arbitrary numbers, enables the compression of the difference between the final weights and the helper string, instead of the weights themselves. 
The usage of a helper string yields tighter generalization bounds, such as in \cite{ComputingNonvacuous}.

\begin{defn}
Let $G_{\mathcal{W},s} = \{ g_{W,s} | W \in \mathcal{W} \}$ be a class of classifiers with trainable parameters $W$ and a helper string $s$. A classifier $f$ is $(\gamma, \mathbf{s})$-compressible with respect to $G_{\mathcal{W},s}$ if there exists $W \in \mathcal{W}$ such that for any sample in the dataset $x \in \mathbf{s}$, we have for all $k$
\begin{align}
|f(x)[k] - g_{W,s}(x)[k] | \leq \gamma.
\end{align}
\end{defn}
%Note that $y$ is the correct label vector of the input $x$, with $k^*$ the correct class of the input $x$, i.e. $y[k^*] = 1$, $y[k]=0$ $\forall k \neq k^*$.
Note that $f(x)[k]$ is the $k$-th entry in the $K$-dimensional vector $f(x)$.
The aforementioned definition leads to the following theorem for a general classifier.

\begin{thm}
Suppose $G_{\mathcal{W},s} = \{ g_{W,s} | W \in \mathcal{W} \}$, where $W$ is a set of $q$ parameters, each of which can have at most $r$ discrete values, and $s$ is a helper string. Let $\mathbf{s}$ be a training set with $N$ samples. If the trained classifier $f$ is $(\gamma,\mathbf{s})$-compressible with  $G_{\mathcal{W},s}$, then there exists $W \in \mathcal{W}$ for which with high probability over the training set,
\begin{align}
{\ell}_{exp,0}{(g_{W,s})} \leq {\ell}_{emp,\gamma}(f,\mathbf{s}) + O \left( \sqrt{\frac{q \log r}{N}} \right).
\end{align}
\end{thm}
This theorem formalizes the generalization abilities of the compression of a classifier $f$.
Relying on this finding, a compression algorithm which yields a bounded generalization error on the output of deep neural networks is proposed in \cite{StrongerBoundsCompression}. This compression algorithm changes the weights of the neural network using a variation of matrix projection.
We refer the reader to \cite{StrongerBoundsCompression} for the corresponding bound for CNNs, along with the empirical findings that establish that it is tighter than the familiar ones based on the product of the weight matrix norms, which are shown to be quite loose.

Another recent work \cite{Compressibility_and_Generalization} also takes a compression-based approach to examining the generalization of DNNs, and provides some interesting complementary insights. A generalization bound for compressed networks based on their compressed size is given, and it is shown that the compressibility of models that tend to over-fit is limited, meaning more bits would be necessary to save a trained network which over-fits its training dataset.

\subsection{Train faster, generalize better: stability of stochastic gradient descent}
\label{Train faster, generalize better: Stability of stochastic gradient descent}

The approach of examining generalization through the lens of the commonly used stochastic gradient optimization methods is taken in \cite{Train_Faster_Generalize_Better}.
It is essentially claimed that any model trained with a stochastic gradient method for a reasonable amount of time would exhibit a small generalization error.

Much insight is given into why the usage of stochastic gradient methods yields good generalization in practice, along with a formal foundation as to why popular techniques for training deep neural networks promote the stability of the obtained solution.
It is argued that stochastic gradient methods are useful in achieving a low generalization error since as long as the objective function is smooth and the number of taken steps is sufficiently small these methods are stable.
Relying on the definitions and bounds for algorithm stability from \cite{Stability_and_Generalization}, stability bounds for both convex and non-convex optimization problems are derived under standard Lipschitz and smoothness assumptions.

An interesting aspect is the relation between an algorithm's generalization error and the amount of training epochs used during its optimization process. When an algorithm is trained for an arbitrarily long training time, it could achieve a small training error by memorizing the training dataset, yet with no generalization abilities. However, an algorithm's ability to fit the training data \emph{rapidly}, with a reasonably small amount of training iterations, is indicative of its ability to generalize well.

It is shown that stochastic gradient methods are uniformly stable. In the convex case, the stability measure decreases as a function of the sum of the optimization step sizes, meaning that these methods reach a solution that generalizes well as long as the optimization steps are sufficiently small and the number of iterations is not too large.
Moreover, for strictly convex loss functions, these methods are stable for an arbitrarily long training time.
Relating to the non-convex case of neural networks, it is shown that the number of training steps of stochastic gradient methods can grow as fast as $N^c$ for $N$ training samples and a small $c>1$, and good generalization would still be achieved. This sheds light on the superior generalization abilities of neural networks, which are trained with many optimization steps.

The following theorem gives a bound for convex loss minimization with a stochastic gradient method.
\begin{thm}
Let the loss function $\ell$ be $\epsilon$-smooth, convex and $M$-Lipschitz. Then a stochastic gradient method with step sizes $\alpha_t \leq \frac{2}{\epsilon}$ for $T$ steps satisfies uniform stability with
\begin{align}
\beta_{stability} \leq \frac{2M^2}{N} \sum_{t=1}^T \alpha_t.
\end{align}
\end{thm}
We leave the formal proof of this theorem to \cite{Train_Faster_Generalize_Better}. We refer the reader to the corresponding stability bound for the non-convex case given in subsection~\ref{Stability and Generalization}.

Moreover, as long as the number of training iterations is linear in the number of data points in the training set, the generalization error is bounded by a vanishing function of the sample size. This means that a short training time by itself can be sufficient to prevent over-fitting, even for models with a large amount of trainable parameters and no explicit regularization.

In addition, a theoretical affirmation is given to the familiar role of regularization in reducing over-fitting and improving the generalization capabilities of learning algorithms. The advantages of using methods such as weight decay regularization, gradient clipping, dropout, projection etc., are formulated and explained.

For instance, the popular technique of dropout decreases the effective Lipschitz constant of the objective function, thus decreasing the bound on the generalization error, as formalized in the following theorem.
\begin{thm}
A randomized map $D:\Omega \rightarrow \Omega$ is a dropout operator with rate $r$ if for every $v \in D$ it holds that $\mathbb{E} \{ ||Dv||_2 \} = r||v||_2$.
For a differentiable function $f:\Omega \rightarrow \Omega$, which is $M$-Lipschitz, the dropout gradient update defined by $\alpha D(\nabla f(v))$, with learning rate $\alpha$ is $(r\alpha M)$-bounded.
\end{thm}
\noindent \textbf{Proof.} Since $f$ is assumed to be differentiable and $M$-Lipschitz, using the linearity of the expectation operator we get that
\begin{align}
\mathbb{E} \{ ||\alpha D(\nabla f(v))|| \} = \alpha r \mathbb{E} || \nabla f(v) || \leq \alpha r M.
\end{align}
This obtained upper bound to the gradient update implies an enhanced stability of the learning algorithm according to the dependency on the Lipschitz constant $M$, which appears in various generalization bounds.

Many other works analyze the characteristics of the loss function and the SGD optimization algorithm used for the training of DNNs as well.
A recent work \cite{SGD_separable_data} shows that even without explicit regularization, for linearly separable logistic regression problems the SGD algorithm converges to the same direction as the max-margin solution, i.e. the solution of the hard margin SVM.

Another recent work \cite{SGD_Overparameterized_linearly_separable} studies the problem of two-layer neural networks with ReLU or Leaky ReLU activations when the data is linearly separable. In the specific examined setting the parameters of the first layer are updated whereas the parameters of the second layer are fixed throughout the training process. Convergence rates of the SGD algorithm to a global minimum are introduced and generalization guarantees for this minimum, which are independent of the network size, are given as well.

Another related work \cite{Multi_Branch_Architectures} examines why DNN architectures that have multiple branches (e.g. Inception, SqueezeNet, Wide ResNet and more) exhibit improved performance in many applications. It is claimed that one cause for this phenomenon is the fact that multi-branch architectures are less non-convex in terms of the duality gap of the optimization problem in comparison to other commonly used DNN architectures. This may explain why the usage of stochastic gradient methods yields improved generalization results for these networks, as it may contribute to their improved stability.

\subsection{On large-batch training for deep learning: generalization gap and sharp minima}
\label{On large-batch training for deep learning: generalization gap and sharp minima}

In \cite{Large_Batch_Training} another point of view on stochastic gradient methods is taken through the examination of the effect of the size of the training mini-batch on the generalization capabilities of the obtained solution.
Though this point of view is mostly empirical, it offers thought-provoking explanations to an interesting phenomenon.

SGD based algorithms perform an optimization step using the gradient of the objective function which is computed on a subset of the training dataset, commonly referred to as a training "mini-batch".
In deep learning, typical mini-batch sizes for training are between several tens to several hundreds of training samples per mini-batch.
It has been empirically observed that training using a larger mini-batch, i.e. more training samples are used to make an optimization step in each iteration, leads to a larger generalization error of the obtained solution.

In \cite{Large_Batch_Training} an explanation to this phenomenon is given by the notion that the usage of large mini-batches encourages convergence to sharp minima solutions to the optimization problem of the training of DNNs (i.e. minimizers of the training loss function), thus obtaining worse generalization. Contrastingly, the usage of small mini-batches tends to lead to solutions with flat minima which yield better generalization.
For the exact definition of the term \emph{sharpness} in this context see Definition~\ref{Sharpness_definition}.

Much light is shed on the aforementioned phenomenon by examining the sharpness of the obtained solutions (minimizers).
It is empirically shown that using large mini-batches during training leads to convergence to a solution with large sharpness, whereas training with small mini-batches leads to a solution with small sharpness (large flatness), which has been linked to better generalization.
The value of the minimum itself (the value of the objective function at the minimizer) in both cases is often very similar, despite the difference in sharpness.

One explanation is that using smaller mini-batches for gradient based steps is essentially equivalent to using a noisy approximation of the gradient, a property that generally leads to convergence to a flatter minimum.
Other conjectures claim that large mini-batches encourage over-fitting, are attracted to saddle points, or simply lack the "explorative" characteristic of small mini-batches.

It is shown in \cite{Large_Batch_Training} that the usage of large mini-batches leads to convergence to sharp minimizers of the objective function, i.e. the Hessian matrix $\nabla^2 \ell_{emp}(\mathcal{A}_{\mathbf{s}_N})$ has a significant amount of large positive eigenvalues, whereas small mini-batches lead to convergence to flat minimizers, with many small eigenvalues in $\nabla^2 \ell_{emp}(\mathcal{A}_{\mathbf{s}_N})$. This leads to the conclusion that large mini-batches effectively prevent the stochastic gradient optimization method from evading undesired basins of attractions.

For the empirical analysis of this phenomenon, an analysis of the Hessian matrix $\nabla^2 \ell_{emp}(\mathcal{A}_{\mathbf{s}_N})$ is required. Due to the significant computational overhead of computing the Hessian matrix in deep learning models, an alternative sharpness measure is employed. We again refer the reader to Definition~\ref{Sharpness_definition} in the preliminaries for the specifics.

Using this metric of sharpness, it is shown on 6 different networks that there is a strong correlation between the usage of small training mini-batches and flatness (i.e. low sharpness) of the obtained solution which leads to a small generalization error (empirically evaluated by the difference between the test and training errors).
Returning to the notion of the usage of small mini-batches being equivalent to noisy approximation of the gradient of the loss function, it is presumed that during training this noise effectively encourages the objective function to exit basins of attraction of sharp solutions towards the ones belonging to flat solutions. Contrastingly, large mini-batches do not have sufficient noise in the gradient to escape sharp minimizers, thus leading to convergence to a worse solution in the sense of a larger generalization error.

Training using a larger mini-batch is highly beneficial, as it allows an increased parallelization of the performed computations, and faster training as a result. For this reason, several mitigation methods for this phenomenon, that will allow the usage of larger mini-batches during training without compromising the generalization capabilities of the obtained solution, are presented.
These methods improve the performance of solutions which are obtained using large mini-batch training. Such methods are data augmentation, conservative training, robust training and others. However, these methods exhibit a limited influence on the sharpness of the attained solution and thereby a limited influence on the generalization error.

\subsection{Sharp minima solutions to the training of DNNs can generalize for deep nets}
\label{Sharp Minima Can Generalize For Deep Nets}

In \cite{SharpMinima} the notion that the flatness of the minima of the loss function obtained using SGD-based optimization algorithms is key in achieving good generalization is examined.
The relation between the geometry of the loss function in the environment of a solution and the obtained generalization error is examined, and through the exploration of different definitions of "flatness" substantial insight is provided into the conjecture that flat minima lead to better generalization.  

With contrast to other prominent works, it is argued that most notions of flatness cannot be directly used to explain generalization.
It is specifically shown that for DNNs with ReLU activations it is possible to apply model re-parameterization and obtain arbitrarily sharper minima.
This essentially means that the notion of flatness can be abstract, and different interpretations of it could lead to very different conclusions.
The reason for this contradiction stems from Bayesian arguments regarding the KL-divergence, which are used to explain the superior generalization ability of flat minima.
Since the KL-divergence is invariant to parameter change, and the notion of flatness is not characterized by such invariance, arguments of flatness can be mistakenly made when more context regarding the definition of flatness is absent.

In this aspect, the work in \cite{SharpMinima} nicely exhibits that even though empirical evidence points to a correlation between flat minima and good generalization, the exact definition of "flatness" in this context is important, as different definitions can lead to very different results and subsequent conclusions.

Several related properties of the Hessian of deep neural networks, their generalization capabilities and the role of the SGD-based optimization are also examined in \cite{Poggio2017TheoryOD}.

\subsection{Train longer, generalize better: closing the generalization gap in large batch training of neural networks}
\label{Train longer, generalize better: closing the generalization gap in large batch training of neural networks}

In an attempt to tackle the phenomenon of degraded generalization error when large mini-batch training is used, a different theoretical explanation, along with a consequent technique to overcome the phenomenon, is suggested in \cite{Train_longer_Generalize_better}.
It is shown that by adjusting the learning rate and using batch normalization during training the generalization gap between small and large mini-batches can be significantly decreased. In addition, it is claimed that there is no actual generalization gap between these two cases; large mini-batch training can generalize just as well as small mini-batch training by adapting the number of training iterations to the mini-batch size.

This claim relies on the conjecture that the initial training phase of a neural network using a stochastic gradient method can be described as a high dimensional "random walk on a random potential" process with "ultra-slow" logarithmic increase in the distance of the weights from their initialization values.
Empirical results show that small mini-batch training produces network weights that are further away from their initial values, compared to the case of large mini-batch training. 
Consequently, by adjusting the learning rate and adding batch normalization to the training algorithm, the generalization gap between small and large mini-batch training can be substantially decreased.
This also implies that the initial training phase with a high learning rate is crucial in obtaining good generalization.
Training longer in the initial high learning rate regime enables the model to reach farther environments in the objective function space, which may explain why it allows the optimization algorithm to find a flatter minima which is correlated with better generalization.

This leads to the conclusion that there is no inherent generalization gap in this case: adapting the amount of training iterations can mitigate the generalization gap between small and large mini-batch training.
Based on these findings, the "Ghost Batch Normalization" algorithm for training using large mini-batches is presented.

\begin{alg} (Ghost Batch Normalization)
\label{Ghost_Batch_Normalization}
\\
\underline{Inputs:} activation values $x$ over a large mini-batch $B_{large} = \{x_1,\dots,x_m \}$ of size $|B_{large}|$, size of virtual small mini-batch $|B_{small}|$ (where $|B_{small}| < |B_{large}|$).
$\gamma, \beta, \eta$ are learned algorithm parameters, where $\eta$ represents the learning momentum.
\\
\underline{Training Phase:} \\
Scatter $B_{large}$ s.t. $\{ X^1, X^2, \dots, X^{\frac{|B_{large}|}{|B_{small}|}} \} = \{x_{1, \dots, |B_{small}|}, x_{|B_{small}|+1, \dots, 2|B_{small}|}, \dots, x_{|B_{large}|-|B_{small}|,\dots,m}\}$ \\
$\mu_B^l \leftarrow \frac{1}{|B_{small}|} \sum_{i=1}^{|B_{small}|} X_i^l$ $\hspace{0.2cm}$ for $l=1,2,3,\dots$ (ghost mini-batch means) \\
$\sigma_B^l \leftarrow \sqrt{\frac{1}{|B_{small}|} \sum_{i=1}^{|B_{small}|} \left( X_i^l - \mu_B^l \right)^2 + \epsilon}$ $\hspace{0.2cm}$ for $l=1,2,3,\dots$ (ghost mini-batch std's) \\
$\mu_{run} = (1-\eta)^{|B_{small}|} \cdot \mu_{run} + \sum_{i=1}^{\frac{|B_{large}|}{|B_{small}|}} (1-\eta)^i \cdot \eta \cdot \mu_B^l$ \\
$\sigma_{run} = (1-\eta)^{|B_{small}|} \cdot \sigma_{run} + \sum_{i=1}^{\frac{|B_{large}|}{|B_{small}|}} (1-\eta)^i \cdot \eta \cdot \sigma_B^l$ \\
return $\gamma \cdot \frac{X^l-\mu_B^l}{\sigma_B^l} + \beta$ \\
\underline{Test Phase:} \\
return $\gamma \cdot \frac{X^l-\mu_{run}^l}{\sigma_{run}^l} + \beta$ $\hspace{0.2cm}$ (scale \& shift)
\end{alg}
This algorithm enables a decrease in the generalization error without increasing the overall number of parameter updates as it acquires the necessary statistics on small virtual ("ghost") mini-batches instead of the original larger mini-batches.

In addition, common practice instructs that during training, when the test error plateaus, one should decrease the learning rate or stop training all together to avoid over-fitting. However, in \cite{Train_longer_Generalize_better} it has been empirically observed that continuing to train, even when the training error decreases and the test error stays roughly the same, results in a test error decrease at a later stage of training when the learning rate is decreased, which is indicative of better generalization.

These results provided the incentive to make the relation to the mini-batch size in \cite{Train_longer_Generalize_better}, supporting the idea that the problem is not in the mini-batch size but rather in the number of training updates.
By prolonging the training time for larger mini-batch training by a factor of $\frac{|B_{large}|}{|B_{small}|}e$, where $|B_{large}|$ and $|B_{small}|$ are the sizes of the large and small training mini-batches respectively and $e$ is the amount of training epochs in the original regime, it is empirically shown how the generalization gap between the two cases can be completely closed.

\subsection{Generalization error of invariant classifiers}
\label{Generalization Error of Invariant Classifiers}

In \cite{Invariant_Classifiers} the generalization error of invariant classifiers is studied. A common case in the field of computer vision is the one in which the classification task is invariant to certain transformations of the input such as viewpoint, illumination variation, rotation etc.
The definition of an invariant algorithm is as follows.
\begin{defn} (Invariant Algorithm)
A learning algorithm $\mathcal{A}$ is invariant to the set of transformations $\mathcal{T}$ if its embedding is invariant:
\begin{align}
f(t_i(x),\mathbf{s}_N) = f(t_j(x),\mathbf{s}_N) \hspace{0.2cm} \forall x \in \mathcal{X}, t_i, t_j \in \mathcal{T}.
\end{align}
where $\mathcal{X}$ is the algorithm's input space and $t_i, t_j$ are transformations.
\end{defn}

It is shown that invariant classifiers may have a much smaller generalization error than non-invariant classifiers, and a relation to the size of the set of transformations that a learning algorithm is invariant to is made.
Namely, it is shown that given a learning method invariant to a set of transformations of size $T$, the generalization error of this method may be up to a factor $\sqrt{T}$ smaller than the generalization error of a non-invariant learning method.
We leave the details of the proof to \cite{Invariant_Classifiers}.

Many other works examine invariant classifiers, as utilizing the property of invariance can lead to improved algorithm performance.
Another prominent work that examines invariant image representations for the purpose of classification is   \cite{Invariant_Scattering_Convolution_Networks}.

\subsection{Generalization error and adversarial attacks}
\label{Generalization Error and Adversarial Attacks}

It has lately been shown that even though deep neural networks typically obtain a low generalization error, and therefore perform their designated tasks with high accuracy, they are highly susceptible to adversarial attacks \cite{Intriguing,Goodfellow15Explaining}, with similarity to other machine learning algorithms. An adversarial attack is a perturbation in the model's input which results in its failure. Adversarial attacks have been shown to be very effective: even when the change in the input is very small they are likely to fool the model, and are usually unnoticeable to the human eye. On top of that, very little knowledge of the attacked network is necessary for an efficient attack to be crafted, and once an adversarial example is obtained it is highly transferable, meaning it is very likely to fool other DNNs as well.

The existence of adversarial attacks exposes an inherent fault in DNN models and their ability to generalize well: although DNNs can generalize very well, they can be very easily fooled.
One should keep in mind that this fault is not unique to deep neural networks and characterizes other machine learning models as well.

In \cite{Sokolic17Robust} a new regularization technique is suggested using the regularization of the Frobenius norm of a network's Jacobian matrix. It is shown that bounding the Frobenius norm of the network's Jacobian matrix reduces the obtained generalization error.
In \cite{Sensitivity_and_Generalization_Google} it is shown that neural networks are more robust to input perturbations in the vicinity of the training data manifold, as measured by the norm of the network's Jacobian matrix.
The correlation between the aforementioned robustness and the network's generalization capabilities is also noted. 
In \cite{ImprovingDnnRobustnessJacobian} this notion is taken further. It is shown that this Jacobian regularization also improves the robustness of DNNs to adversarial attacks, thus showing that reducing a network's generalization error has also collateral benefits.
In \cite{Adversarially_Robust_Generalization} it is shown that the sample complexity (the number of training samples necessary to learn the classification function) of robust learning can be significantly larger than that of standard learning.

A comprehensive survey on the threat of adversarial attacks on deep learning models is given in \cite{AdverarialAttacksSurvey}.

\section{Open problems}
\label{Open Problems}
Given the above overview of generalization error in deep learning, we provide here a list of open problems we have identified that we believe will have an important future impact on the field.

\subsection{Problem 1: Generalization and memorization}
\label{Problem 1: Generalization and Memorization}

As reviewed in subsection~\ref{Understanding deep learning requires rethinking generalization}, understanding the capabilities and method of operation of deep neural networks requires a deeper understanding of the interplay between memorization and generalization.
It has been shown that DNNs are powerful enough to memorize a random training dataset, yet with no actual generalization. It would be expected from a model that over-fits any training data so well to obtain poor generalization, yet in practice DNNs generalize very well. It follows that currently existing theories are lacking since they are unable to explain this phenomenon, and a new comprehensive theory is required.
Furthermore, an algorithm's ability to obtain a low generalization error strongly depends on the provided training dataset and not just on the model architecture. In order for effective learning to take place, the training dataset must be sufficiently large and well spread over the sample space in order to avoid the \emph{curse of dimensionality}, a term widely used to refer to the need of a large amount of training data in high-dimensional problems.
It follows that obtaining prior knowledge on the training dataset or the test dataset e.g. the distributions from which they were drawn, would be highly beneficial in obtaining better generalization.

Several works make a relation between the generalization capabilities of DNNs and the underlying data model. For example, \cite{DNN_Curse_of_Dimensionality} examines which architecture is better for learning different functions. It is shown that deep neural networks, as opposed to shallow networks, are guaranteed to avoid the \emph{curse of dimensionality} for an important class of problems: when the learned function is compositional. A thorough review of the abilities of shallow and deep neural networks to learn different kinds of compositional functions is done.
Another recent work \cite{DNN_VS_KNN} examines the relationship between the classification performed by DNNs and the $K$-NN algorithm applied at the embedding space of these networks. The results suggest that a DNN generalizes by learning a new metric space adapted to the structure of the training dataset.

Following this track, we believe that better generalization bounds for learning algorithms are obtainable when an assumption on the data model is made. For instance, a sparsity assumption on the data model may be useful in this context, as can be observed in several prominent works such as 
\cite{Vainsencher11Sample,Jung14Performance,Gribonval15Sparse,Gribonval15Sample,Schnass18Convergence,Singh18Reconstruction,Papyan17Convolutional}.

\subsection{Problem 2: Generalization and robustness}
\label{Problem 2: Generalization and Robustness}
Another prominent and interesting open problem is understanding the robustness properties of DNNs.
A deep neural network is trained on a specific training dataset, which is sampled from some probability distribution. A good model, which has been adequately trained on a sufficiently large and balanced training dataset, is expected to generalize well on unseen test data which is drawn from the same distribution.

However, the question arises - how well would this DNN generalize to test data which was drawn from a different distribution? Can constraints on the relationship between the training distribution and the test distribution be imposed to guarantee good generalization?
An example for this problem can be taken from the field of computer vision. Let us assume a DNN is trained to classify images which were taken of a scene in daylight. How well would this DNN generalize for images of the same scene taken at night? This is an important problem with implications for numerous applications, for example autonomous vehicles. Networks trained to recognize issues on roads in sunny Silicon Valley may not work well in rainy and foggy London.

Another related and interesting question is how good is the cross-task adaptation of deep neural networks.
How well can a DNN, which was trained to perform a certain task, perform another task? How beneficial would incremental learning \cite{Incremental_Learning} (a continuous "online" training of the model with new data) and transfer learning \cite{Transfer_Learning} (a wide term which in the implementational case is typically used to refer to the the training of several layers of a network, which have been previously trained for performing a different task or on a different training dataset, for the purpose of performing a new task or to work well with a different data source) be in this case? How adaptive is a neural network between different tasks, and are there key design guidelines for obtaining better generalization, robustness and transferability for a network? How is the necessary information for good generalization embedded in the learned features, and in what sense is the network essentially learning a new metric?
We refer the reader to \cite{TransferLearningSurvey} for a comprehensive survey of the field of transfer learning.

All of these questions represent highly important areas of research with substantial significance to the design of better DNN architectures with better generalization, robustness and transferability.

\subsection{Problem 3: Generalization and adversarial examples}
\label{Problem 3: Generalization and Adversarial Examples}

A special case of the previous problem is the one of adversarial examples, which we have presented in subsection~\ref{Generalization Error and Adversarial Attacks}. The counter-intuitive vulnerability of DNNs to adversarial examples opens the door to a new angle in the research of the generalization of deep neural networks. It is of high importance to have a comprehensive theory dedicated to this type of examples. Some theories for explaining this phenomenon have been suggested, such as in \cite{Goodfellow15Explaining,Intriguing,Transferable}. However, this still remains an active field of research.

\subsection{Problem 4: Generalization error of generative models}
\label{Problem 4: Generalization beyond classification}

An important lens through which the generalization capabilities of DNNs is examined is that of generative models.
Generative models are models which are used to learn the underlying distribution from which the data is drawn, and thus manufacture more data from the same probability distribution, which can be used for training.

For example, Generative Adversarial Networks (GANs) \cite{GANs} are a model which consists of two networks, a generative network $G$ that captures the data distribution, and a discriminative network $D$ that estimates the probability of a training sample being either genuine or fake (i.e. manufactured by $G$). In this minimax problem, the training phase results in $G$ learning the underlying distribution of the training data. 
The work in \cite{Arora17Generalization} provided initial results regarding the generalization properties of GANs. 

Another generative model which has gained much traction in the past years is the Variational Auto-Encoder (VAE) \cite{VAEs}. Classical VAEs are based on two neural networks, an encoder network and a decoder network. The encoder is used to learn a \emph{latent variable}, from which the decoder generates a sample which is similar to the original input to the encoder, i.e. which is approximately drawn from the same distribution.

The question arises - which probability distributions can be learned under which conditions? How beneficial to the generalization capabilities would training on additional manufactured data be? These questions represent additional substantial paths for research with a promising impact on the field of deep learning.

\subsection{Problem 5: Generalization error and the information bottleneck}
\label{Problem 5: Generalization error and the information bottleneck}

Recently, the information bottleneck has been introduced to explain generalization and convergence in deep neural networks \cite{DL_Information_Bottleneck,Opening_the_box_DL_information}. By characterizing the DNN \textit{Information Plane} -- the plane of the mutual information values that each layer preserves on the input and output variables -- it is suggested that a network attempts to optimize the information bottleneck trade-off between compression and prediction for each layer.

It is also known that the information bottleneck problem is related to the information-theoretic noisy lossy source coding problem (a variation of the lossy source coding problem) \cite{Elements_of_information_theory}.
In particular, a noisy lossy source coding problem with a specific loss function gives rise to the information bottleneck.
Therefore, it is of interest to explore links between information theory, representation learning and the information bottleneck, in order to cast insights onto the performance of deep neural networks under an information-theoretic lens.
Preliminary steps in this direction are taken in \cite{compression_multi_task_learning,Information_bottleneck_representation_learning}.

\section{Conclusions}
\label{Conclusions}

Even though deep neural networks were shown to be a promising and powerful machine leaning tool which is highly useful in many tasks, the source of their capabilities remains somewhat elusive.
Deep learning models are highly expressive, over-parameterized, complex, non-convex models, which are usually trained (optimized) with a stochastic gradient method.

In this article we reviewed the generalization capabilities of these models, shedding light on the reasons for their ability to generalize well from the training phase to the test phase, thus maintaining a low generalization error. We reviewed some of the fundamental work on this subject and also provided some more recent findings and theoretical explanations to the generalization characteristics of deep neural networks and the influence of different parameters on their performance.

We also reviewed various emerging open problems in deep learning, ranging from the interplay between robustness, generalization and memorization, to robustness to adversarial attacks, the generalization error of generative models and the relation between the generalization error and the information bottleneck.
These open problems require a deeper understanding to fully unlock the potential applicability of deep learning models in real environments.
Beyond these open problems there are various other interesting learning settings that require much additional theoretical research, such as multi-modal learning, multi-task learning, incremental learning, the capacity of neural networks, the optimization of deep networks and more.
While the current state of research in this field is promising, we believe that much room remains for further work to provide more comprehensive theories and a better understanding of this important subject as outlined throughout this article.

\bibliographystyle{splncs}
\bibliography{references}

\end{document}